%% file: arxiv_RCD.tex
\documentclass[10pt,twocolumn,letterpaper]{article}

\usepackage[pagenumbers]{wacv}

\usepackage{algorithm}
\usepackage{algpseudocode}
\usepackage{booktabs}
\usepackage{siunitx}
\usepackage{multirow}
\usepackage{dblfloatfix}
\usepackage{tabularx}
\usepackage[table]{xcolor}
\usepackage{makecell}
\usepackage{tocloft}
\usepackage[most]{tcolorbox}

\definecolor{wacvblue}{rgb}{0.21,0.49,0.74}
\usepackage[pagebackref,breaklinks,colorlinks,allcolors=wacvblue]{hyperref}
\usepackage{etoc} 
\hypersetup{
  pdftitle={Resolve, Clear, or Defer: Training-Free Active Prototype Triage for Ecological Detection Pipelines},
  pdfauthor={Anirudh Potlapally, Brijesh Nanda Brunda Yogananda, Jianyang Gu, Maanav Jajodia, Luke Meyers, Wei-Lun Chao, Hari Subramoni}
}

\definecolor{defaultrow}{gray}{0.93}

\def\wacvPaperID{1315}
\def\confName{WACV}
\def\confYear{2027}

\title{Resolve, Clear, or Defer: Training-Free Active Prototype Triage for
Ecological Detection Pipelines}

\author{
Anirudh Potlapally$^{1}$ \quad
Brijesh Nanda Brunda Yogananda$^{1}$ \quad
Jianyang Gu$^{1}$ \quad
Maanav Jajodia$^{1}$ \\
Luke Meyers$^{2}$ \quad
Wei-Lun Chao$^{3}$ \quad
Hari Subramoni$^{1}$ \\[1mm]
$^{1}$The Ohio State University \quad
$^{2}$University of Puerto Rico, R\'io Piedras \quad
$^{3}$Boston University \\
{\tt\small
\{potlapally.2,brundayogananda.1,gu.1220,jajodia.6,subramoni.1\}@osu.edu} \\
{\tt\small luke.meyers@upr.edu \quad chao209@bu.edu}
}

\begin{document}
\maketitle

\input{sec/0_abstract}
\input{sec/1_introduction}
\input{sec/2_related_work}
\input{sec/3_training_free_active_triage}
\input{sec/4_experiments_and_results}
\input{sec/5_conclusion}

{\small
  \bibliographystyle{ieeenat_fullname}
  \bibliography{main}
}

\clearpage

\onecolumn
\thispagestyle{empty}
\pdfbookmark[0]{Supplementary Material}{supplementary}
\null\vspace*{\fill}
\begin{center}
    {\LARGE\bfseries Supplementary Material:\par}
    \vspace{0.8em}
    {\LARGE\bfseries
    Resolve, Clear, or Defer: Training-Free Active Prototype Triage for
    Ecological Detection Pipelines\par}
\end{center}
\vspace*{\fill}\null
\clearpage
\twocolumn

\setcounter{section}{0}
\setcounter{subsection}{0}
\setcounter{subsubsection}{0}
\setcounter{figure}{0}
\setcounter{table}{0}
\setcounter{equation}{0}
\setcounter{algorithm}{0}
\setcounter{footnote}{0}
\renewcommand{\theHsection}{supp.\arabic{section}}
\renewcommand{\theHsubsection}{supp.\arabic{section}.\arabic{subsection}}
\renewcommand{\theHsubsubsection}{supp.\arabic{section}.\arabic{subsection}.\arabic{subsubsection}}
\renewcommand{\theHfigure}{supp.\arabic{figure}}
\renewcommand{\theHtable}{supp.\arabic{table}}
\renewcommand{\theHequation}{supp.\arabic{equation}}
\providecommand{\theHalgorithm}{}
\renewcommand{\theHalgorithm}{supp.\arabic{algorithm}}

\etocsetlocaltop.toc{part}
\begingroup
\setcounter{tocdepth}{2} 
\footnotesize
\setlength{\parskip}{0pt}
\setlength{\cftbeforetoctitleskip}{0pt}
\setlength{\cftaftertoctitleskip}{5pt}
\setlength{\cftbeforesecskip}{2pt}
\setlength{\cftbeforesubsecskip}{0pt}
\setlength{\cftsecindent}{0em}
\setlength{\cftsubsecindent}{1.5em}
\renewcommand{\cfttoctitlefont}{\bfseries\normalsize}
\begin{tcolorbox}[
    enhanced,breakable,colback=white,colframe=black!35,
    boxrule=0.5pt,arc=2pt,left=7pt,right=7pt,top=5pt,bottom=5pt
]
\localtableofcontents
\end{tcolorbox}
\endgroup
\vspace{0.5em}

\input{supp/0_overview}
\input{supp/1_method_details}
\input{supp/2_evaluation_protocol}

\input{supp/3_baseline_details}

\input{supp/4_expanded_results}

\input{supp/5_sensitivity}

\input{supp/6_runtime}

\input{supp/7_qualitative}

\end{document}

%% file: sec/0_abstract.tex
\begin{abstract}
Ecological monitoring increasingly relies on camera traps and aerial surveys that generate large image collections. High-recall animal detectors can localize candidate animals at scale. However, they also produce many detections that still require expert review, including false positives, ambiguous cases, and animals difficult to identify at the species level. Since exhaustive review requires substantial effort, we propose training-free active prototype triage, where reviewed detections become
reusable evidence to (1) \emph{Resolve} supported foreground classes,
(2) \emph{Clear} likely background false positives, or
(3) \emph{Defer} insufficiently supported cases for expert review. Across camera-trap and aerial datasets, foreground recognition and false-positive rejection benefit from different pretrained representations: BioCLIP2 provides foreground evidence while DINOv3 provides background evidence, reducing erroneous foreground clearing while improving automatic foreground recovery. Deferring unsupported cases avoids the higher error rates caused by forced decisions. 
Compared with confidence-thresholded multiclass $k$-NN and
uncertainty-sampling logistic regression, prototype triage improves
foreground recovery while maintaining competitive or lower review burden. Acquisition goes beyond class discovery: even after classes are represented, prototype choice determines how reliably evidence propagates through the remaining detections. Our results suggest that once animal localization reaches high recall, the challenge shifts from finding animals to deciding which examples to review, how to reuse them, and when evidence does not support an automatic decision.
\end{abstract}

%% file: sec/1_introduction.tex
\section{Introduction}
\label{sec:introduction}

Camera traps and aerial surveys increasingly generate image collections that
are difficult to review exhaustively. High-recall class-agnostic animal
detectors and open-vocabulary localization models can localize animals
without deployment-specific training
\cite{beery2019efficientpipelinecameratrap,liu2024groundingdinomarryingdino,OWLv2_2023,
carion2026sam3segmentconcepts}. Localization enables object-level analysis,
which can improve recognition by preserving animal detail and reducing background context \cite{to_crop_or_not_2024}, but substantially
increases the number of instances requiring review. These detections may include
background false positives, partial or ambiguous animals, and cases difficult
to identify at the species level; ecological datasets are often
imbalanced, with rare species represented by a few examples
\cite{norouzzadeh2021active,bothmann2023automated,kurinchivendhan2026needles}.
\textbf{The bottleneck therefore shifts from finding animals to deciding which
detections merit expert attention and how reviewed examples can be reused to
handle the rest.} Table~\ref{tab:proposal_generator_pr_iou} illustrates this regime across camera-trap and aerial imagery.

Pretrained encoders allow proposal crops to be compared without updating the
underlying representation
\cite{Stevens_2024_CVPR,gu2025bioclip,simeoni2025dinov3}, enabling
expert-labeled examples to serve directly as nonparametric evidence
\cite{Zhang_2022_ECCV}. Active learning provides mechanisms for selecting
proposals for expert review
\cite{sener2018activelearningconvolutionalneural,safaei2025active}, while
prototype-based recognition and selective prediction provide complementary
mechanisms for reusing labeled support and abstaining on unsupported cases
\cite{snell2017prototypicalnetworksfewshotlearning,narasimhan2024plugin}.
What remains less explored is an iterative fixed-pool setting that couples
selection, reuse, and deferral: \textbf{each new expert label should immediately alter
which remaining proposals can be recognized, cleared as background, or retained
for further review, without parameter updates.}

\begin{figure*}[t]
    \centering
    \includegraphics[width=0.8\textwidth]{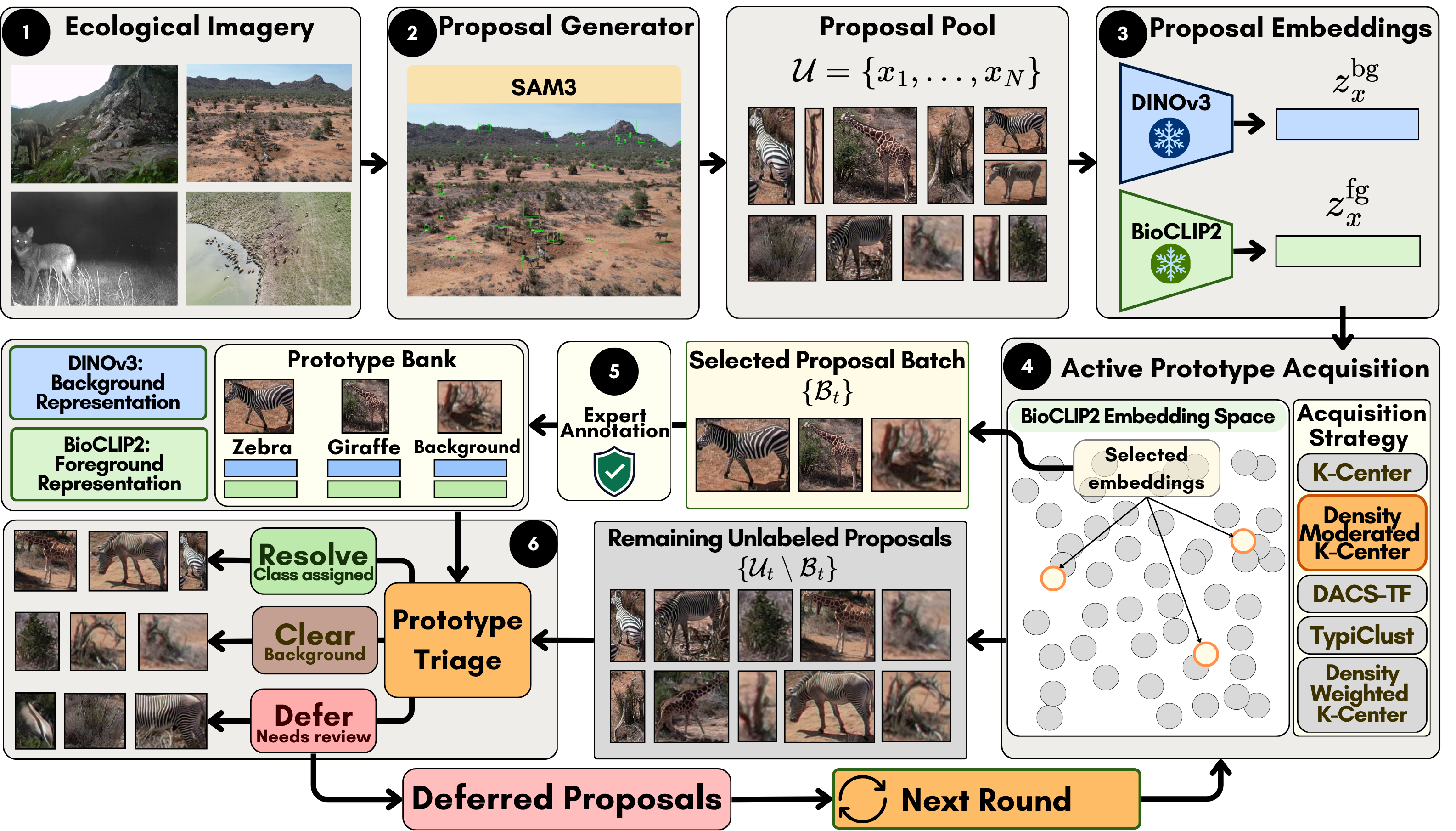}
    \caption{\textbf{Overview of the training-free active prototype triage framework.}
    SAM3 generates animal proposals embedded with frozen BioCLIP2 \cite{gu2025bioclip} and DINOv3 \cite{simeoni2025dinov3}
    encoders. Acquisition selects proposals for expert review and prototype-bank expansion. For every unreviewed proposal, BioCLIP2 provides foreground semantic evidence
    and DINOv3 provides background evidence. Proposals are \emph{Resolved}, \emph{Cleared}, or \emph{Deferred}; \emph{Deferred}
    proposals carry into the next round. No detector, encoder, or classifier is
    trained or fine-tuned.}
    \label{fig:overview}
    \vspace{-1.5em}
\end{figure*}

We propose a \emph{training-free active prototype triage}
framework\footnote{\url{https://github.com/OSU-SAI-Lab/ProtoRCD}}
(Figure~\ref{fig:overview}) built around an evolving
prototype bank. Each round selects unresolved proposals for expert review and
adds their frozen representations as verified prototypes. The
proposals are re-triaged as \emph{Resolve}, \emph{Clear}, or \emph{Defer},
allowing each annotation to immediately influence the rest of the pool without
updating the detector, encoders, or a downstream classifier. This setting raises three coupled questions. First, which proposals should be
reviewed so that the prototype bank captures useful variation rather than
redundant or isolated examples? Second, should foreground recognition and
background clearing rely on the same pretrained representation, or do these
roles benefit from different evidence? Third, when is prototype support
too weak to justify an automatic decision, making \emph{Defer} preferable to
forcing \emph{Resolve} or \emph{Clear}?

\input{tables/table_proposal_generator}

Across two camera-trap and two aerial datasets, we find that
\textbf{foreground recognition and background clearing benefit from different
pretrained representations}: BioCLIP2 \cite{gu2025bioclip} foreground evidence combined with DINOv3 \cite{simeoni2025dinov3}
background evidence lowers mean erroneous foreground clearing while increasing foreground recovery across all four datasets. \textbf{Deferral is equally
important}: forcing unsupported cases into automatic decisions increases both
wrong-class and foreground-clearing errors in all five seeds. Compared with confidence-thresholded multiclass $k$-NN and
uncertainty-sampling logistic regression, prototype triage improves
foreground recovery while maintaining competitive or lower review burden. We further find that \textbf{acquisition goes beyond class discovery}:
even after classes are represented, the examples selected for review
determine how reliably their support propagates through the remaining
pool. While no strategy dominates every objective, our proposed density-moderated
$k$-center (DM) achieves competitive results at low computational cost
by balancing geometric coverage with local proposal density.

%% file: tables/table_proposal_generator.tex
\begin{table}[t]
\centering
\captionsetup{skip=2pt}
\caption{\textbf{Proposal-generator comparison.}
Precision (P) and recall (R) are macro-averaged within each modality. SAM3 maintains
high recall across camera-trap and aerial imagery and is
therefore used as the common proposal generator. MegaDetector v6 is included
as a camera-trap-oriented reference.}
\label{tab:proposal_generator_pr_iou}
\footnotesize
\setlength{\tabcolsep}{4pt}
\begin{tabular}{llcccc}
\toprule
Modality & Detector & P@.25 & R@.25 & P@.50 & R@.50 \\
\midrule
Camera trap & SAM3 & 82.1 & 98.7 & 80.1 & 96.3 \\
Camera trap & MDv6 & 90.1 & 96.8 & 89.1 & 95.7 \\
Aerial & SAM3 & 73.6 & 97.4 & 71.1 & 94.2 \\
Aerial & MDv6 & 87.3 & 47.9 & 82.0 & 44.9 \\
\bottomrule
\end{tabular}
\vspace{-1.8em}
\end{table}

%% file: sec/2_related_work.tex
\section{Related Work}
\label{sec:related_work}

\noindent\textbf{Ecological vision and frozen representations.}
Ecological vision pipelines separate animal localization from downstream
classification. Camera-trap recognition is challenging under environmental
variation: models that perform well at training locations can generalize
poorly to new camera-trap sites \cite{beery2018recognition}, while detection
in complex forest scenes must contend with small animals, camouflage, and
cluttered backgrounds \cite{song2024benchmarking}. Aerial wildlife detection
faces related challenges including small objects, class imbalance, limited
annotations, and complex backgrounds \cite{xu2024aerialreview}.
MegaDetector detects generic ``animal'' regions for classification
\cite{beery2019efficientpipelinecameratrap}, while models such as
Grounding DINO, OWLv2, and SAM3 provide flexible open-vocabulary localization
\cite{liu2024groundingdinomarryingdino,OWLv2_2023,carion2026sam3segmentconcepts}.
After detection, proposal instances are cropped or focused for classification
\cite{to_crop_or_not_2024}, and frozen encoders such as BioCLIP2, TaxaBind,
CATALOG, and DINOv3 provide general-purpose or biologically specialized frozen
representations for these tasks
\cite{Stevens_2024_CVPR,gu2025bioclip,sastry2025taxabind,
Santamaria_2025_WACV,simeoni2025dinov3}. Ecological end users often configure
these methods with low thresholds for high recall
\cite{to_crop_or_not_2024}, as false positives can be reviewed, but missed
animals are difficult to recover, potentially affecting the biological
interpretation of results. These tools support high-recall objectives but do
not provide ways to process the large, noisy detection pools they generate.

\noindent\textbf{Nonparametric recognition and evolving support.}
Few-shot and cache-based methods reuse labeled examples, prototypes, or stored
features for nonparametric or training-free inference
\cite{vinyals2016matchingnetworksshotlearning,
snell2017prototypicalnetworksfewshotlearning,wang2019simpleshot,
Zhang_2022_ECCV,Karmanov_2024_CVPR}. Biological benchmarks have likewise
evaluated $k$-NN and nearest-centroid inference over frozen representations
\cite{picek2025fungitasticmultimodaldatasetbenchmark}. Unlike methods that rely on a pre-annotated reference set, we acquire prototypes through iterative expert review of the target proposal pool.

\noindent\textbf{Active learning, transductive labeling, and selective prediction.}
Pool-based active learning uses coverage, typicality, representativeness, and
density-aware criteria to select examples for annotation
\cite{sener2018activelearningconvolutionalneural,
hacohen2022activelearningbudgetopposite,Submodularity_facility_2015,DACS_2022},
with extensions to vision--language models and human interaction
\cite{safaei2025active,Beck_2024_WACV}. FLAME starts from high-recall
open-vocabulary proposals and actively acquired annotations, but trains a
lightweight classifier for adaptation \cite{Refael_2026_WACV}; our reviewed
proposals instead form an evolving nonparametric memory that repeatedly
triages the remaining pool without parameter updates.

Ecological active learning has addressed species identification and counting
\cite{norouzzadeh2021active,bothmann2023automated}, while transductive
work treats ecological annotation as fixed-pool completion with rare-class
discovery and stopping criteria \cite{kurinchivendhan2026needles}. Selective
prediction and open-set active annotation study abstention and unknown samples
\cite{wu2024confidence,narasimhan2024plugin,narasimhan2024longtail,
Ning_2022_CVPR,Zong_2025_CVPR}. We combine these perspectives in an iterative
training-free workflow where reviewed examples provide evolving evidence for
\emph{Resolve}, \emph{Clear}, and \emph{Defer} decisions over the remaining
proposal pool.

%% file: sec/3_training_free_active_triage.tex
\section{Training-Free Active Prototype Triage of Object Proposals}
\label{sec:method}

Consider an ecological survey in which a high-recall detector produces many
object regions, or proposals. Rather than reviewing every proposal, we label a
small batch whose examples become prototypes for the rest. Supported animals
are assigned a foreground class (\emph{Resolve}), likely false positives are
removed (\emph{Clear}), and insufficiently supported cases remain for review
(\emph{Defer}). Repeating this process expands the prototype bank without
retraining any model.

\subsection{Detection-to-Triage Formulation}
\label{sec:problem_workflow}

Let $\mathcal{U}=\{x_i\}_{i=1}^{N}$ denote a fixed proposal pool with unknown
labels $y_i\in\{0,1,\ldots,C\}$, where $0$ is background and $1,\ldots,C$ are
foreground animal classes. Before active review, we precompute
$\ell_2$-normalized foreground and background representations
$\mathbf{z}^{\mathrm{fg}}_x$ and $\mathbf{z}^{\mathrm{bg}}_x$ for each proposal.
We use the dual-encoder configuration, with BioCLIP2~\cite{gu2025bioclip} providing the foreground representation and DINOv3-Huge+~\cite{simeoni2025dinov3} the background
representation.

Starting from an empty prototype bank, each round selects unresolved proposals
for expert review and adds their labels and frozen representations to the bank.
The remaining unresolved proposals are then re-triaged. Reviewed and
automatically \emph{Resolved} or \emph{Cleared} proposals leave the unresolved
pool, but only reviewed proposals enter the prototype bank. \emph{Deferred}
proposals remain for subsequent rounds until the review budget is exhausted or
the pool is empty. Each reviewed prototype therefore consists of an expert
label together with its foreground and background representations.

\subsection{Prototype-Based Resolve, Clear, and Defer}
\label{sec:triage_rule}

Each unresolved proposal is compared with the same verified prototype bank in
separate foreground and background representation spaces. For a calibrated
neighborhood size $k$, both voting tests use the strict two-thirds
supermajority
\begin{equation}
\tau_{\mathrm{fg}}=\tau_{\mathrm{bg}}
=
\left\lfloor \frac{2k}{3}\right\rfloor+1.
\label{eq:vote_threshold}
\end{equation}
Let $\mathcal{N}^{\mathrm{fg}}_k(x)$ and
$\mathcal{N}^{\mathrm{bg}}_k(x)$ contain up to $k$ nearest verified prototypes
in each space. If fewer than $k$ verified prototypes are
available, all available prototypes are used, but the vote threshold remains
as defined by the calibrated $k$.

Among the foreground neighbors, foreground-labeled prototypes vote by class.
Let $\hat{c}(x)$ be the unique class receiving the most votes and
$n_{\mathrm{fg}}(x)$ its vote count. If no unique winner exists,
$F(x)=\mathrm{false}$. We set $\rho_c=+\infty$ for classes with fewer than two
prototypes. A foreground winner is accepted only with sufficient
votes and similarity:

\begin{equation}
F(x)=
\left[n_{\mathrm{fg}}(x)\ge\tau_{\mathrm{fg}}\right]
\wedge
\left[s_{\mathrm{fg}}(x,\hat{c})\ge\rho_{\hat{c}}\right].
\end{equation}
where
\begin{equation}
s_{\mathrm{fg}}(x,c)=
\max_{p\in\mathcal{P}_c}
\cos\!\left(
\mathbf{z}^{\mathrm{fg}}_x,
\mathbf{z}^{\mathrm{fg}}_p
\right),
\end{equation}
$\mathcal{P}_c$ is the prototype set for class $c$, and $\rho_c$ is
its class-specific similarity floor. If $F(x)$ holds, $x$ is
\emph{Resolved} to $\hat{c}(x)$.

Let $n_{\mathrm{bg}}(x)$ be the number of background-labeled prototypes in
$\mathcal{N}^{\mathrm{bg}}_k(x)$. If $F(x)$ is false and
\begin{equation}
n_{\mathrm{bg}}(x)\ge\tau_{\mathrm{bg}},
\label{eq:background_predicate}
\end{equation}
$x$ is \emph{Cleared}. Otherwise it is \emph{Deferred}. \emph{Resolve} takes
precedence over \emph{Clear}.

\textbf{Class-specific similarity floor.}
For classes with at least two prototypes, we measure how similar their
prototypes are to one another. For every $p\in\mathcal{P}_c$, we find its most
similar other prototype from the same class, and $\rho_c$ is the lowest of
these similarities. A proposal
can satisfy the similarity condition for class $c$ only if its similarity to
at least one prototype in $\mathcal{P}_c$ reaches this within-class floor. We
recompute $\rho_c$ after each acquisition round as the prototype bank grows.
At least $\tau_{\mathrm{fg}}$ reviewed prototypes from class $c$ are necessary,
but not sufficient, for the foreground voting condition to be satisfied.

\textbf{Calibration.}
The neighborhood size $k$ is selected before active review from a calibration
bank excluded from acquisition and evaluation. The selected $k$ and
corresponding vote thresholds remain fixed during active review, while the
class-specific floors $\rho_c$ are recomputed as the prototype bank grows.
Calibration and annotation accounting are detailed in
Sec.~\ref{sec:experimental_protocol}. The complete iterative
acquisition-and-triage procedure is provided in the supplementary material.

\subsection{Active Prototype Acquisition}
\label{sec:acquisition}

Acquisition selects which unresolved proposals receive expert labels and enter
the prototype bank. Selection is performed in the foreground representation
space (BioCLIP2). We compare training-free strategies spanning random sampling,
geometric coverage, typicality, representativeness, and density-aware coverage.

\textbf{Acquisition baselines.}
We compare Random, $k$-center~\cite{sener2018activelearningconvolutionalneural},
TypiClust~\cite{hacohen2022activelearningbudgetopposite}, Sampled Facility
Location (Facility)~\cite{Submodularity_facility_2015}, and DACS-TF~\cite{DACS_2022}.
Random samples unresolved proposals uniformly. $k$-center uses farthest-first
traversal. TypiClust selects typical samples from underrepresented clusters,
while Facility greedily maximizes representativeness. DACS-TF adapts
Density-Aware Core-Set~\cite{DACS_2022} using density-stratified groupwise
$k$-center. Implementation details are provided in the supplement.

\textbf{Density-aware $k$-center.}
Prior work has explored density-aware coverage-based sampling
\cite{climaco2026densityawarefarthestpointsampling,
yan2026beyond}.
We incorporate local density into $k$-center selection by scoring each
proposal $i$ as
\begin{equation}
a_i=d(i,\mathcal{S})\,w(r_i),
\label{eq:weighted_kcenter}
\end{equation}
where $d(i,\mathcal{S})$ is the cosine distance to the nearest proposal
already selected in batch $\mathcal{S}$ and $r_i\in[0,1]$ is its
percentile-ranked local density. We consider
\begin{equation}
w(r)=
\begin{cases}
1,       & \text{$k$-center},\\
r,       & \text{Density-weighted (DW)},\\
r(1-r),  & \text{Density-moderated (DM)}.
\end{cases}
\label{eq:density_weights}
\end{equation}
DW favors dense regions while retaining the farthest-first coverage objective.
Pure coverage can over-select isolated proposals, whereas monotonic density
weighting can favor dominant modes. For prototype acquisition, we instead seek
representative but nonredundant examples: very sparse proposals may provide
weak support for propagation, while the densest regions may contain many
similar examples. We therefore introduce DM, whose weight is largest at
intermediate density and decreases toward both extremes.

Local density is computed once from nearest-neighbor cosine similarity over the
initial acquisition pool. Batchwise $k$-center is initialized with the proposal
nearest the unresolved-pool centroid, whereas weighted variants begin with the
proposal maximizing $w(r_i)$.

\input{tables/table_dataset_stats}

%% file: tables/table_dataset_stats.tex
\begin{table}[t]
\centering
\caption{\textbf{Object proposal pools.}
Image and proposal counts are reported after dataset selection, proposal
assignment, and filtering. ``FG cls.'' denotes the number of foreground
classes, ``FG prev.'' foreground prevalence, and ``Imbal.'' the
largest-to-smallest foreground-class count ratio. Counts include the
calibration bank.}
\label{tab:datasets}

\footnotesize

\setlength{\tabcolsep}{1.4pt}

\begin{tabular}{llrrrrr}
\toprule
Dataset & Modality & Images & Proposals & FG cls. & FG prev. & Imbal. \\
\midrule
ENA24       & Camera Trap     & 8,650  & 11,165 & 21 & 86.4\% & 21$\times$ \\
LoTE-Animal & Camera Trap     & 19,787 & 39,690 & 11 & 79.6\% & 39$\times$ \\
WAUD        & Aerial & 651    & 19,840 & 3  & 83.1\% & 11$\times$ \\
MMLA-Mpala  & Aerial & 1,540  & 29,150 & 2  & 65.9\% & 5$\times$ \\
\bottomrule
\end{tabular}
\vspace{-1em}
\end{table}

%% file: sec/4_experiments_and_results.tex
\section{Experimental Setup}
\label{sec:experimental_setup}

\begin{figure}[t]
    \centering
    \includegraphics[width=\columnwidth]{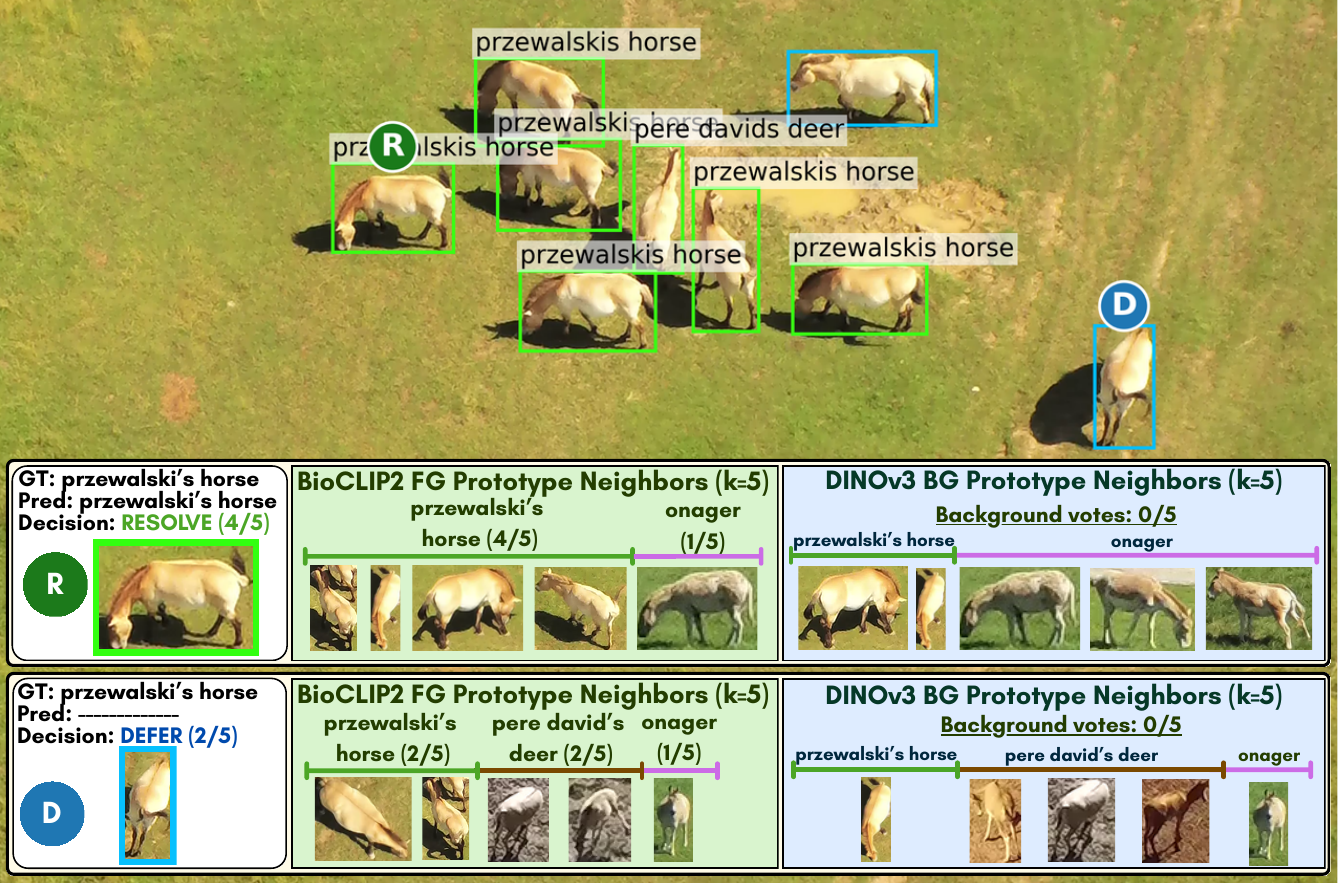}
    
    \caption{
    \textbf{Qualitative prototype-voting example on WAUD.}
    At round 3 under DM $k$-center acquisition, two Przewalski's horse proposals
    from the same scene receive different triage decisions.
    The resolved proposal receives four of five BioCLIP2 prototype votes for
    Przewalski's horse, whereas the deferred proposal receives mixed support across
    Przewalski's horse, P\`ere David's deer, and onager, with two of five
    neighbors supporting the GT class. Both examples
    receive zero DINOv3 background votes and therefore do not trigger \emph{Clear}.
    }
    \label{fig:qualitative_analysis}
    \vspace{-1.75em}
\end{figure}

\subsection{Datasets and Proposal Generation}
\label{sec:datasets}

We evaluate on four ecological proposal pools spanning camera-trap and aerial
imagery and covering multiclass recognition, class imbalance, fine-grained
appearance, and dense proposal regimes. Table~\ref{tab:datasets} summarizes
the dataset statistics.

\textbf{ENA24.}
ENA24~\cite{yousif2019dynamic} provides a broad multiclass camera-trap setting
with 21 foreground classes, requiring reviewed support to propagate across
many species.

\textbf{LoTE-Animal.}
LoTE-Animal~\cite{Liu_2023_ICCV} provides a strongly imbalanced multiclass
camera-trap setting in which frequent classes dominate the proposal pool.
Because our evaluation is transductive over a fixed pool, we combine its train
and test splits. Label access follows the protocol in
Sec.~\ref{sec:experimental_protocol}.

\textbf{WAUD.}
Wilds Aerial Ungulate Detection (WAUD) is a three-class aerial pool containing
P\`ere David's deer, Przewalski's horse, and onager. We construct it by
combining onager imagery and annotations from MMLA-Wilds
\cite{kline2025mmla,mmla_wilds,kline2025wildwing} with SmartWilds imagery
\cite{kline2025smartwildsmultimodalwildlifemonitoring,thewilds_drones_2025},
for which we manually annotate P\`ere David's deer and Przewalski's horse.
The visually similar ungulates make WAUD a fine-grained recognition setting
under class imbalance.

\textbf{MMLA-Mpala.}
We use a multi-video subset of MMLA-Mpala
\cite{kline2025mmla_dataset,kline2025mmla,KABR_Raw_Videos} containing
zebra--giraffe imagery with dense backgrounds. This provides an aerial,
small-object, low-cardinality, and proposal-dense setting for evaluating
prototype propagation.

Full class distributions and dataset-selection details are provided in the
supplementary material.

\textbf{Proposal generation.}
We generate proposals with SAM3~\cite{carion2026sam3segmentconcepts} using the
text prompt \emph{animal}. We retain detections with score $\geq 0.4$, apply
non-maximum suppression at IoU~0.5, and use SAHI-style tiled inference
\cite{sahi_2022} when needed for small animals.
Table~\ref{tab:proposal_generator_pr_iou} shows that SAM3 maintains high recall
across camera-trap and aerial imagery and is therefore used as the common
proposal generator.

Each proposal is assigned the class of its highest-overlap ground-truth box at
IoU~$\geq 0.25$ and is otherwise treated as background. The lower assignment
threshold reduces sensitivity to box disagreement for small, occluded, or
partially visible animals. For WAUD, ground-truth boxes and proposals with
minimum side length below 16 pixels are excluded because these instances lack
sufficient visual detail for reliable species-level recognition.

\subsection{Evaluation Metrics}
\label{sec:metrics}

The goal of triage is not simply to maximize automatic decisions. It must
reduce remaining expert effort while recovering foreground animals, preserving
class-balanced recognition, and avoiding erroneous removal of true animals as
background. We therefore use three primary metrics and two targeted error
diagnostics.

Let $N$ denote the size of the calibration-excluded evaluation pool, $H$ the
number of proposals actively reviewed, $D$ the number remaining
\emph{Deferred} at the endpoint, and $N_{\mathrm{FG}}$ the number of
true-foreground proposals.

\textbf{Review Burden}
measures the fraction of the evaluation pool that has been actively reviewed or
still requires review:
\[
    \mathrm{Review\ Burden} = \frac{H + D}{N}.
\]
This proposal-count measure captures endpoint annotation demand rather than
annotation time or cognitive effort. The separate calibration bank is excluded
from $N$ and represents an additional one-time calibration cost.

\textbf{Automatic FG Recall}
measures the fraction of true-foreground proposals correctly
\emph{Resolved} to their foreground class without expert review:
\[
    \mathrm{Automatic\ FG\ Recall}
    =
    \frac{\#\{\text{correct automatic FG Resolves}\}}
         {N_{\mathrm{FG}}}.
\]
Foreground proposals that are reviewed, \emph{Deferred}, \emph{Cleared}, or
\emph{Resolved} to the wrong foreground class remain in the denominator but do
not count as recovered.

\textbf{Conditional FG Macro-F1}
measures class-balanced recognition among true-foreground proposals
automatically \emph{Resolved} to a foreground class. The macro average uses all
foreground classes in the evaluation pool together with any additional
foreground classes predicted automatically. Classes with no eligible automatic
foreground decision contribute zero. Expert-reviewed, \emph{Deferred}, and
\emph{Cleared} foreground proposals are excluded. If no eligible automatic
foreground decisions exist, the metric is undefined.

Two additional diagnostics measure automatic-decision reliability.
\textbf{False Clear Rate} is the fraction of true-foreground proposals
incorrectly \emph{Cleared} as background:
\[
    \mathrm{False\ Clear\ Rate}
    =
    \frac{\#\{\text{true FG proposals Cleared as BG}\}}
         {N_{\mathrm{FG}}}.
\]
\textbf{Automatic Error Rate} is the fraction of automatic
\emph{Resolve}/\emph{Clear} decisions that are incorrect, including wrong
foreground-class resolutions, foreground proposals incorrectly cleared as
background, and background proposals incorrectly resolved as foreground. We
use this metric only when evaluating explicit \emph{Defer}.

For acquisition trajectories, we additionally plot
\textbf{Deferred (\%)} at each review budget, defined as the fraction of the
calibration-excluded evaluation pool still \emph{Deferred}. This differs from
endpoint Review Burden because the horizontal axis already records the number
of proposals reviewed.

\subsection{Experimental Protocol}
\label{sec:experimental_protocol}

Prototype-triage experiments review 50 proposals per round for at most
16 rounds, corresponding to at most 800 active-review labels, and stop early
when no Deferred proposals remain. Results are averaged over five random seeds.

Before active review, we draw a disjoint 100-proposal calibration bank for each dataset and seed. Its labels are used only for parameter selection. Calibration examples are excluded from acquisition, evaluation, and the active prototype bank, and represent a separate one-time expert cost.

We select the neighborhood size
$k \in \{5,7,9,11\}$ by leave-one-out triage on the calibration bank,
prioritizing fewer erroneous foreground clears, more correct automatic
decisions, and fewer deferrals. Ties favor smaller $k$. The selected value is
fixed during active review, and vote thresholds follow
Eq.~\ref{eq:vote_threshold}. Per-seed calibrated values and the full selection
rule are provided in the supplementary material.

We use 30 neighbors for density estimation and 20 for TypiClust
typicality~\cite{hacohen2022activelearningbudgetopposite}. Unless otherwise
stated, every proposal is embedded by frozen BioCLIP2 and DINOv3-Huge+.
BioCLIP2 provides foreground-class evidence and the acquisition
representation, while DINOv3 provides background evidence. The encoders have
distinct evidence roles for every proposal rather than routing different
proposals through separate branches.

\section{Results and Discussion}
\label{sec:results}

\subsection{Representation Choice and Abstention}
\label{sec:routing_defer}

\input{tables/table_representation_routing}

\textbf{Do foreground recognition and background clearing benefit from
different representations?}
Table~\ref{tab:routing_ablation} compares foreground/background encoder
pairs. Using BioCLIP2 for both roles (BioCLIP2/BioCLIP2) yields strong
Conditional FG Macro-F1, while replacing only the background encoder
with DINOv3 (BioCLIP2/DINOv3) lowers False Clear Rate and increases
Automatic FG Recall on all four pools.

These gains also depend on which encoder serves each role.
On LoTE-Animal, reversing the pairing (DINOv3/BioCLIP2) raises
False Clear Rate from $4.1\%$ to $13.2\%$ and reduces Automatic FG
Recall from $74.4\%$ to $61.4\%$. This is consistent with BioCLIP2's
biologically specialized representations supporting class-specific
foreground evidence and DINOv3's more general visual features
supporting heterogeneous background evidence. We therefore use
BioCLIP2 for foreground evidence and acquisition and DINOv3 for
background evidence in the remaining analyses.

\input{tables/table_forced_decision}

\textbf{Is explicit deferral necessary?}
Table~\ref{tab:defer_forced} tests whether \emph{Deferred} proposals can be
converted into automatic decisions without sacrificing reliability. Forcing
decisions lowers Review Burden and increases Automatic FG Recall, but
raises Automatic Error Rate and False Clear Rate. ENA24's False Clear Rate
more than triples from $2.3\%$ to $7.3\%$.

The higher error rates under forced assignment indicate that \emph{Deferred} proposals are less reliably classified using the current prototype bank.

\begin{figure}[t]
    \centering
    \includegraphics[width=1.0\linewidth]{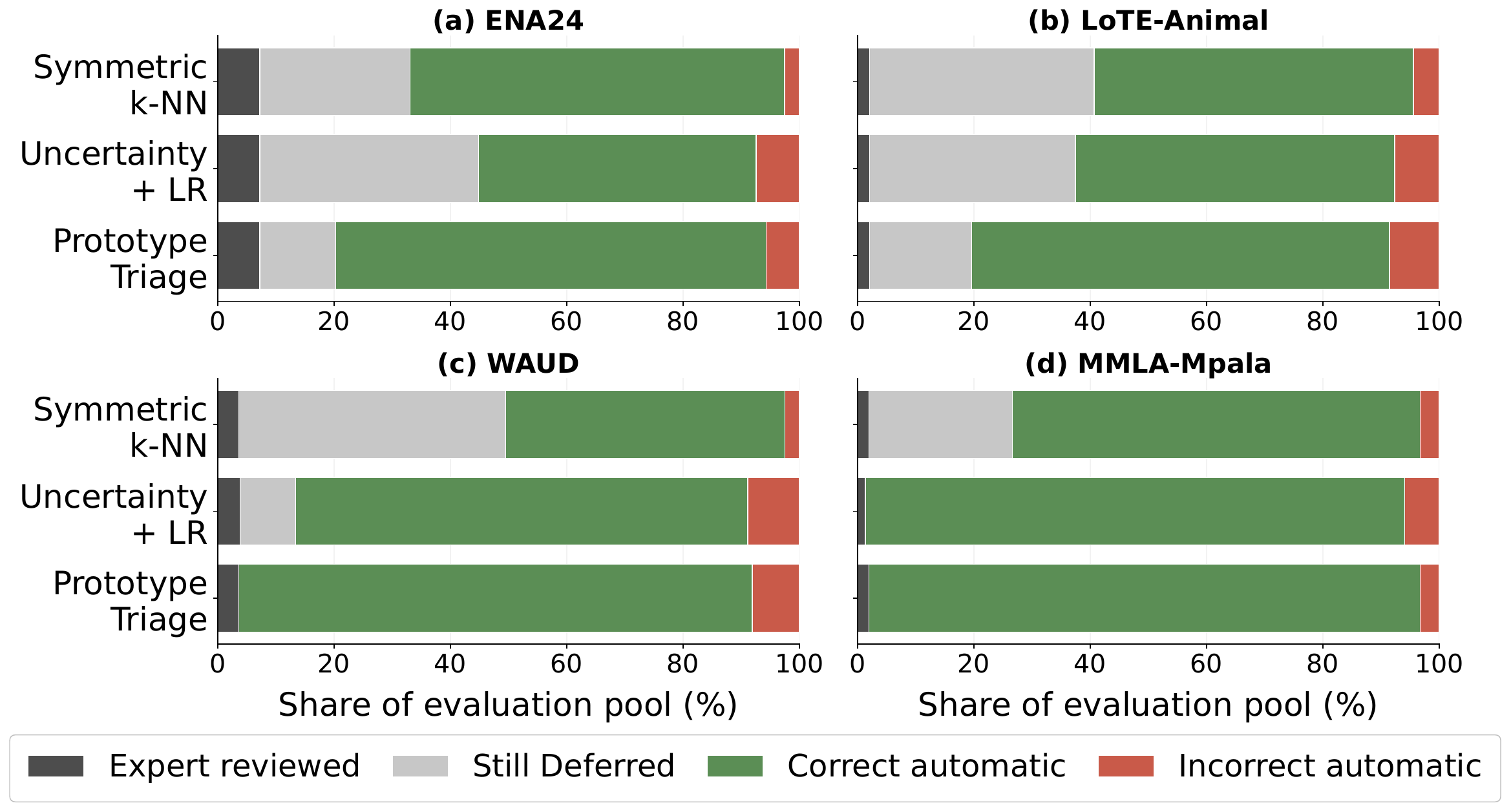}
    \caption{\textbf{Endpoint workload decomposition.}
Each bar partitions the calibration-excluded evaluation pool into
expert-reviewed, still \emph{Deferred}, correct automatic, and incorrect
automatic proposals. Expert-reviewed and still-\emph{Deferred} proposals
together constitute Review Burden. Prototype Triage shifts a larger share of
the pool to correct automatic decisions while reducing Review Burden on ENA24,
LoTE-Animal, and WAUD. On MMLA-Mpala, LR attains slightly lower Review Burden,
while Prototype Triage produces more correct automatic decisions.}
    \label{fig:workload_decomposition}
    \vspace{-1.8em}
\end{figure}

\subsection{Comparison with Selective Classification Baselines}
\label{sec:selective_baselines}

\textbf{Can conventional selective or fitted alternatives recover the same
workload--recovery operating point?}
We compare Prototype Triage with a symmetric nonparametric baseline and
a fitted active-learning alternative under the same maximum active-review
budget.

\textbf{Symmetric $k$-NN.}
This baseline reuses the same DM-acquired reviewed examples as Prototype
Triage at each round. The $k$ nearest reviewed samples in BioCLIP2 space vote
over foreground classes and background, with the winning vote fraction
determining prediction or abstention. Background is treated symmetrically with
foreground, without separate \emph{Resolve}/\emph{Clear} tests,
class-specific foreground constraints, or role-specific representations.

\textbf{Uncertainty + logistic regression (LR).}
This fitted alternative follows its own active-review trajectory in BioCLIP2
space. LR is refit after each round, predictive entropy selects the next review
batch, and the top-class posterior controls abstention. Full calibration and
implementation details for both baselines are provided in the supplementary
material.

\input{tables/table_selective_fitted}

Table~\ref{tab:selective_baselines} shows that Prototype Triage achieves the
highest Automatic FG Recall on all four datasets. The comparison with Symmetric $k$-NN isolates how evidence is reused,
since both methods receive the same reviewed examples. Its higher Review Burden on ENA24,
LoTE-Animal, and WAUD reflects weaker propagation rather than acquisition.
Paired-seed comparisons are reported in the supplementary material. Symmetric
$k$-NN can achieve strong conditional foreground recognition while abstaining
on more proposals. On WAUD, it leaves $45.9\%$ of the pool Deferred, while
Prototype Triage approaches complete pool processing.

These results, along with the representation-role ablation, support separating
class-specific foreground evidence from background evidence rather than relying
on a single symmetric multiclass confidence criterion. Prototype Triage
automates more of the pool while retaining \emph{Defer} for cases without
sufficient prototype evidence.

Uncertainty + LR also trails Prototype Triage in Automatic FG Recall on all
four datasets, although the gap narrows on the lower-cardinality aerial pools.
This is consistent with fitted classifiers becoming more competitive as class
structure simplifies, while local prototype evidence remains useful under
multiclass imbalance. Under controlled cumulative online time after frozen
feature extraction, LR is faster on LoTE-Animal and WAUD, while Prototype
Triage is $2.0$--$6.9\times$ faster than Symmetric $k$-NN (Table~\ref{tab:selective_baselines}).

Prototype Triage provides a consistent balance between expert workload,
foreground recovery, and online cost while retaining explicit
\emph{Resolve}/\emph{Clear}/\emph{Defer} controls.

\input{tables/table_endpoint_main}

\begin{figure*}[t]
    \centering
    \includegraphics[width=\textwidth]{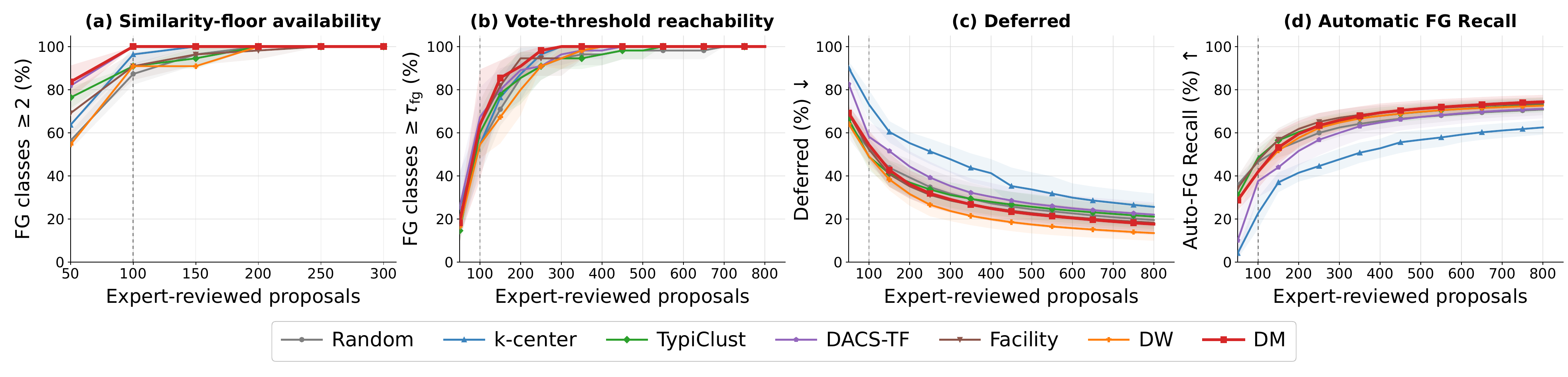}
    \captionsetup{skip=2pt}
    \caption{\textbf{Active acquisition dynamics on long-tailed LoTE-Animal.}
    From left to right: foreground classes with at least two reviewed
    prototypes, foreground classes with enough reviewed support to reach the
    calibrated foreground vote threshold $\tau_{\mathrm{fg}}$, Deferred
    proposals, and Automatic FG Recall as expert review increases.
    The dashed line marks 100 reviewed proposals. Panel (a) is truncated after
    similarity-floor availability saturates. Curves show five-seed means with
    $\pm 1$ standard deviation. Full acquisition trajectories for all datasets
    are provided in the supplementary material.}
    \label{fig:acquisition_dynamics}
    \vspace{-1.5em}
\end{figure*}

\subsection{Active Sampling Strategies}
\label{sec:acquisition_results}

\textbf{How does acquisition shape prototype propagation?}
Figure~\ref{fig:acquisition_dynamics} examines this process on long-tailed
LoTE-Animal. By 100 active-review labels, DM and DACS-TF have at least two
reviewed prototypes for every foreground class, allowing all class-specific
similarity floors to be defined. These similarity floors are recomputed as the
reviewed prototype bank grows, whereas the calibrated foreground vote threshold
$\tau_{\mathrm{fg}}$ remains fixed during active review. The vote-threshold
view is therefore stricter: even after a class has a defined similarity floor,
it needs at least $\tau_{\mathrm{fg}}$ reviewed prototypes before it can
possibly satisfy the foreground voting condition. As review progresses, this
support expands while the Deferred fraction decreases and Automatic FG Recall
increases. The two support views thus distinguish early class coverage from
the stronger support needed for automatic foreground resolution. Full
acquisition and class-support trajectories for all datasets are provided in the
supplementary material.

The aerial datasets provide complementary operating regimes. On WAUD, nearly
all methods reduce Deferred proposals to near zero, while endpoint Automatic FG
Recall ranges from $88.4\%$ for $k$-center to $94.6\%$ for DM, showing that
prototype selection continues to influence foreground recovery even at high
automation. On MMLA-Mpala, the acquisition strategies achieve similarly low
Review Burden and similar foreground recognition, reflecting the greater
redundancy of this low-cardinality pool.

\textbf{What makes a useful acquisition strategy?}
Figure~\ref{fig:acquisition_dynamics} and
Table~\ref{tab:acquisition_endpoints} show that acquisition strategies
differ in their trade-offs between Review Burden, foreground recovery,
and class-balanced recognition. Plain batchwise $k$-center emphasizes
geometric coverage, whereas density-aware strategies also account for
local proposal density. On strongly imbalanced LoTE-Animal, DW achieves
the lowest Review Burden, while DM provides higher Automatic FG Recall
and Conditional FG Macro-F1.

DM favors intermediate-density proposals while retaining the
$k$-center coverage objective. It achieves the highest Automatic FG
Recall on LoTE-Animal and WAUD while maintaining competitive Review
Burden and Conditional FG Macro-F1. Its cumulative acquisition time
remains below one second across all four pools, unlike TypiClust,
Facility, and DACS-TF (see supplementary material).

%% file: tables/table_representation_routing.tex
\begin{table}[t]
\centering
\captionsetup{skip=3pt}
\caption{\textbf{Representation-role ablation.}
Five-seed mean$\pm$std endpoint results after up to 16 rounds
(800 active-review labels). Each configuration is rerun end-to-end,
with acquisition performed in its FG representation space and
$k$ calibrated using the same protocol.
$\star$ marks the configuration used in the remaining experiments.}
\label{tab:routing_ablation}

\footnotesize
\setlength{\tabcolsep}{1.2pt}
\renewcommand{\arraystretch}{0.92}

\resizebox{\columnwidth}{!}{%
\begin{tabular}{@{}lcllcccc@{}}
\toprule
Dataset &
&
\makecell{FG\\encoder} &
\makecell{BG\\encoder} &
\makecell{Review\\Burden $\downarrow$} &
\makecell{Auto-FG\\Recall $\uparrow$} &
\makecell{Cond.\ FG\\Macro-F1 $\uparrow$} &
\makecell{False Clear\\Rate $\downarrow$} \\
\midrule

\multirow{4}{*}{ENA24}
& & DINOv3 & DINOv3
& 24.59$\pm$1.94
& 67.01$\pm$1.31
& .89$\pm$.01
& 2.67$\pm$0.46 \\

& & BioCLIP2 & BioCLIP2
& 22.96$\pm$4.36
& 70.97$\pm$3.22
& \textbf{.96$\pm$.01}
& 3.06$\pm$0.42 \\

& & DINOv3 & BioCLIP2
& 33.34$\pm$8.52
& 58.68$\pm$6.31
& .89$\pm$.01
& 2.59$\pm$1.26 \\

\cmidrule(lr){3-8}
& $\star$ & BioCLIP2 & DINOv3
& \textbf{20.27$\pm$1.19}
& \textbf{74.08$\pm$0.48}
& .94$\pm$.01
& \textbf{2.26$\pm$0.68} \\

\midrule

\multirow{4}{*}{\makecell[c]{LoTE-\\Animal}}
& & DINOv3 & DINOv3
& 18.54$\pm$6.06
& 68.50$\pm$4.06
& .81$\pm$.02
& 5.53$\pm$1.20 \\

& & BioCLIP2 & BioCLIP2
& 23.38$\pm$4.95
& 70.24$\pm$3.30
& \textbf{.95$\pm$.03}
& 5.32$\pm$1.28 \\

& & DINOv3 & BioCLIP2
& \textbf{17.94$\pm$8.97}
& 61.45$\pm$5.35
& .79$\pm$.04
& 13.16$\pm$3.61 \\

\cmidrule(lr){3-8}
& $\star$ & BioCLIP2 & DINOv3
& 19.60$\pm$4.95
& \textbf{74.38$\pm$3.20}
& .92$\pm$.02
& \textbf{4.15$\pm$0.53} \\

\midrule

\multirow{4}{*}{WAUD}
& & DINOv3 & DINOv3
& \textbf{2.92$\pm$0.52}
& 93.20$\pm$1.05
& .94$\pm$.03
& 2.20$\pm$0.56 \\

& & BioCLIP2 & BioCLIP2
& 3.72$\pm$1.30
& 93.97$\pm$1.11
& \textbf{.95$\pm$.01}
& 1.22$\pm$0.57 \\

& & DINOv3 & BioCLIP2
& 3.12$\pm$0.50
& 92.61$\pm$2.16
& .88$\pm$.09
& 1.28$\pm$0.48 \\

\cmidrule(lr){3-8}
& $\star$ & BioCLIP2 & DINOv3
& 3.64$\pm$0.57
& \textbf{94.56$\pm$0.50}
& \textbf{.95$\pm$.02}
& \textbf{1.05$\pm$0.37} \\

\midrule

\multirow{4}{*}{\makecell[c]{MMLA-\\Mpala}}
& & DINOv3 & DINOv3
& 1.14$\pm$0.23
& 95.57$\pm$2.93
& 1.00$\pm$.00
& 3.39$\pm$3.03 \\

& & BioCLIP2 & BioCLIP2
& 1.49$\pm$0.12
& 92.79$\pm$0.63
& 1.00$\pm$.00
& 5.91$\pm$0.62 \\

& & DINOv3 & BioCLIP2
& \textbf{0.72$\pm$0.11}
& 93.64$\pm$0.73
& 1.00$\pm$.00
& 5.73$\pm$0.72 \\

\cmidrule(lr){3-8}
& $\star$ & BioCLIP2 & DINOv3
& 1.93$\pm$0.15
& \textbf{95.58$\pm$0.10}
& 1.00$\pm$.00
& \textbf{2.55$\pm$0.13} \\

\bottomrule
\end{tabular}%
}

\vspace{-1.5em}
\end{table}

%% file: tables/table_forced_decision.tex
\begin{table}[t]
\centering

\caption{\textbf{Effect of explicit Defer.}
Five-seed mean$\pm$std endpoint results after up to 800 active-review
labels. \emph{No Deferral} assigns all remaining Deferred proposals automatically
without changing reviewed support or the prototype bank.}
\label{tab:defer_forced}
\vspace{-2mm}

\scriptsize
\setlength{\tabcolsep}{0.5pt}
\renewcommand{\arraystretch}{0.9}

\resizebox{\columnwidth}{!}{%
\begin{tabular}{@{}llcccc@{}}
\toprule
Dataset & Setting &
\makecell{Review\\Burden $\downarrow$} &
\makecell{Auto-FG\\Recall $\uparrow$} &
\makecell{Auto. Error\\Rate $\downarrow$} &
\makecell{False Clear\\Rate $\downarrow$} \\
\midrule

\multirow{2}{*}{ENA24}
& No Deferral
& \textbf{7.23$\pm$0.00}
& \textbf{79.27$\pm$0.23}
& 14.42$\pm$0.22
& 7.33$\pm$0.32 \\

& With Defer
& 20.27$\pm$1.19
& 74.08$\pm$0.48
& \textbf{7.17$\pm$0.84}
& \textbf{2.26$\pm$0.68} \\

\midrule

\multirow{2}{*}{\makecell[c]{LoTE-\\Animal}}
& No Deferral
& \textbf{2.02$\pm$0.00}
& \textbf{81.55$\pm$0.45}
& 18.52$\pm$0.30
& 9.17$\pm$0.46 \\

& With Defer
& 19.60$\pm$4.95
& 74.38$\pm$3.20
& \textbf{10.54$\pm$1.79}
& \textbf{4.15$\pm$0.53} \\

\bottomrule
\end{tabular}%
}
\end{table}

%% file: tables/table_selective_fitted.tex
\begin{table}[t]
\centering
\caption{\textbf{Baselines under limited expert review.}
Five-seed mean$\pm$std endpoints under at most 800 active-review labels.
Symmetric $k$-NN reuses the DM-acquired reviewed support of Prototype
Triage, while Uncertainty + LR follows its own uncertainty-sampling
trajectory. Review Burden includes proposals already reviewed and those
still Deferred. Online time is cumulative controlled iterative computation
after frozen embeddings, including acquisition-stage, decision, and
per-round recalibration where applicable.}
\label{tab:selective_baselines}

\scriptsize
\setlength{\tabcolsep}{1.7pt}
\renewcommand{\arraystretch}{0.94}

\resizebox{\columnwidth}{!}{%
\begin{tabular}{@{}llcccc@{}}
\toprule
Dataset & Method &
\makecell{Review\\Burden $\downarrow$} &
\makecell{Auto-FG\\Recall $\uparrow$} &
\makecell{Cond.\ FG\\Macro-F1 $\uparrow$} &
\makecell{Online\\Time (s) $\downarrow$} \\
\midrule

\multirow{3}{*}{ENA24}
& Symmetric $k$-NN
& 33.03$\pm$8.04 & 65.13$\pm$7.11 & .97$\pm$.01 & 4.716$\pm$.052 \\
& Uncertainty + LR
& 44.81$\pm$19.20 & 45.25$\pm$14.23 & .75$\pm$.07 & 2.074$\pm$1.460 \\
& Prototype Triage
& \textbf{20.27$\pm$1.19} & \textbf{74.08$\pm$0.48} & .94$\pm$.01 & 1.873$\pm$.037 \\

\midrule
\multirow{3}{*}{\makecell[c]{LoTE-\\Animal}}
& Symmetric $k$-NN
& 40.67$\pm$17.21 & 60.56$\pm$12.05 & \textbf{.96$\pm$.03} & 15.092$\pm$.306 \\
& Uncertainty + LR
& 37.46$\pm$10.51 & 54.72$\pm$6.81 & .81$\pm$.09 & 2.074$\pm$1.538 \\
& Prototype Triage
& \textbf{19.60$\pm$4.95} & \textbf{74.38$\pm$3.20} & .92$\pm$.02 & 7.368$\pm$1.435 \\

\midrule
\multirow{3}{*}{WAUD}
& Symmetric $k$-NN
& 49.50$\pm$29.78 & 47.19$\pm$29.95 & \textbf{.98$\pm$.02} & 5.400$\pm$.828 \\
& Uncertainty + LR
& 13.43$\pm$11.31 & 86.14$\pm$9.48 & .95$\pm$.03 & .147$\pm$.020 \\
& Prototype Triage
& \textbf{3.64$\pm$0.57} & \textbf{94.56$\pm$0.50} & .95$\pm$.02 & .890$\pm$.121 \\

\midrule
\multirow{3}{*}{\makecell[c]{MMLA-\\Mpala}}
& Symmetric $k$-NN
& 26.62$\pm$8.07 & 87.35$\pm$4.41 & 1.00$\pm$.00 & 5.816$\pm$.536 \\
& Uncertainty + LR
& \textbf{1.33$\pm$0.65} & 92.74$\pm$0.40 & 1.00$\pm$.00 & 1.339$\pm$.414 \\
& Prototype Triage
& 1.93$\pm$0.15 & \textbf{95.58$\pm$0.10} & 1.00$\pm$.00 & .840$\pm$.025 \\

\bottomrule
\end{tabular}%
}
\end{table}

%% file: tables/table_endpoint_main.tex
\begin{table}[t]
\centering
\caption{\textbf{Prototype-triage acquisition endpoints.}
Five-seed mean$\pm$std endpoints after at most 16 rounds
(800 active-review labels) using the dual encoder configuration.
Review Burden is the fraction of the calibration-excluded evaluation
pool that has either been actively reviewed or remains Deferred.
No acquisition strategy dominates all objectives.}
\vspace{-2mm}
\label{tab:acquisition_endpoints}
\scriptsize
\setlength{\tabcolsep}{2.5 pt}
\renewcommand{\arraystretch}{0.9}

\resizebox{\columnwidth}{!}{%
\begin{tabular}{llccc}
\toprule
Dataset & Method &
\shortstack{Review\\Burden $\downarrow$} &
\shortstack{Auto-FG\\Recall $\uparrow$} &
\shortstack{Cond.\ FG\\Macro-F1 $\uparrow$} \\
\midrule

\multirow{7}{*}{ENA24}
& Random
& 21.21$\pm$1.68
& 74.00$\pm$0.53
& .93$\pm$.01 \\
& $k$-center
& 40.62$\pm$3.33
& 50.58$\pm$6.76
& .85$\pm$.07 \\
& TypiClust
& 21.30$\pm$1.33
& 75.44$\pm$0.68
& .96$\pm$.00 \\
& Facility
& 19.93$\pm$0.97
& 75.31$\pm$0.39
& .94$\pm$.01 \\
& DACS-TF
& 27.93$\pm$0.41
& 70.94$\pm$0.83
& .96$\pm$.00 \\
& DW
& 19.99$\pm$0.53
& 74.84$\pm$0.45
& .93$\pm$.01 \\
& DM
& 20.27$\pm$1.19
& 74.08$\pm$0.48
& .94$\pm$.01 \\

\midrule

\multirow{7}{*}{\shortstack{LoTE-\\Animal}}
& Random
& 21.51$\pm$5.64
& 70.76$\pm$2.60
& .85$\pm$.04 \\
& $k$-center
& 27.61$\pm$6.19
& 62.48$\pm$3.36
& .85$\pm$.04 \\
& TypiClust
& 23.10$\pm$5.79
& 72.90$\pm$3.36
& .91$\pm$.02 \\
& Facility
& 20.27$\pm$3.87
& 73.59$\pm$2.90
& .89$\pm$.01 \\
& DACS-TF
& 24.02$\pm$5.70
& 71.15$\pm$4.46
& .93$\pm$.01 \\
& DW
& 15.45$\pm$3.57
& 72.59$\pm$2.17
& .87$\pm$.02 \\
& DM
& 19.60$\pm$4.95
& 74.38$\pm$3.20
& .92$\pm$.02 \\

\midrule

\multirow{7}{*}{WAUD}
& Random
& 3.82$\pm$0.57
& 92.60$\pm$1.93
& .91$\pm$.05 \\
& $k$-center
& 2.62$\pm$0.14
& 88.39$\pm$0.47
& .83$\pm$.01 \\
& TypiClust
& 3.95$\pm$0.56
& 92.72$\pm$0.57
& .95$\pm$.01 \\
& Facility
& 4.02$\pm$0.24
& 93.63$\pm$0.40
& .95$\pm$.01 \\
& DACS-TF
& 3.26$\pm$0.30
& 92.88$\pm$0.19
& .91$\pm$.02 \\
& DW
& 4.17$\pm$0.38
& 91.56$\pm$0.63
& .90$\pm$.01 \\
& DM
& 3.64$\pm$0.57
& 94.56$\pm$0.50
& .95$\pm$.02 \\

\midrule

\multirow{7}{*}{\shortstack{MMLA-\\Mpala}}
& Random
& 1.93$\pm$0.12
& 95.70$\pm$0.28
& 1.00$\pm$.00 \\
& $k$-center
& 2.26$\pm$0.10
& 95.92$\pm$0.14
& 1.00$\pm$.00 \\
& TypiClust
& 1.89$\pm$0.10
& 95.86$\pm$0.17
& 1.00$\pm$.00 \\
& Facility
& 1.80$\pm$0.11
& 95.68$\pm$0.17
& 1.00$\pm$.00 \\
& DACS-TF
& 2.25$\pm$0.12
& 96.15$\pm$0.16
& 1.00$\pm$.00 \\
& DW
& 1.78$\pm$0.10
& 95.64$\pm$0.11
& 1.00$\pm$.00 \\
& DM
& 1.93$\pm$0.15
& 95.58$\pm$0.10
& 1.00$\pm$.00 \\

\bottomrule
\end{tabular}%
}

\vspace{-2.5em}
\end{table}

%% file: sec/5_conclusion.tex
\section{Limitations and Future Work}
\label{sec:limitations}

Our evaluation is retrospective and transductive, with expert review
simulated from reference labels. Missing annotations can cause valid animal
proposals to be treated as background. Recovery also depends on upstream
proposal recall and frozen representation quality. The camera-trap pools
contain 11 and 21 foreground classes, compared with two and three in the
aerial pools (Table~\ref{tab:datasets}). Performance differences cannot
be attributed to modality alone because class cardinality and dataset
composition also differ.

Class-specific similarity floors evolve with reviewed support, while
calibrated voting thresholds remain fixed. Classes with fewer than
$\tau_{\mathrm{fg}}$ reviewed prototypes cannot be automatically
\emph{Resolved} and must rely on \emph{Defer}. Adaptive voting and
recalibration under deployment shift remain future work.

Storing BioCLIP2 and DINOv3 embeddings increases storage costs for
growing pools. We measure review workload in proposal counts rather than
annotation time and do not model labeling noise or annotator disagreement.
Reviewed and confidently handled proposals could support semi-supervised
detector training. Streaming data, noisy feedback, and larger domain or
taxonomic shifts remain future work.

\section{Conclusion}

High-recall ecological detectors localize animals at scale but leave many
proposals for expert review. Our training-free active prototype triage
reuses reviewed examples to \emph{Resolve} supported foreground classes,
\emph{Clear} likely background, or \emph{Defer} uncertain cases without
updating detector, encoder, or classifier parameters.

Across camera-trap and aerial pools, separate foreground and background
representations improved automatic foreground recovery and reduced
erroneous clearing. Explicit deferral reduced automatic errors compared
with forced decisions. Matched-support symmetric $k$-NN comparisons
showed that how reviewed evidence is reused matters even when the
reviewed proposals are identical.

Acquisition also influenced recovery beyond class discovery by affecting
how reviewed support propagated through the pool. Density-moderated
$k$-center provided a lightweight, consistently competitive acquisition
strategy across the evaluated datasets.

\section*{Acknowledgments}

This work was supported in part by the NSF AI Institute
for Intelligent Cyberinfrastructure with Computational
Learning in the Environment (ICICLE) under Award
No. OAC-2112606. We gratefully acknowledge computational
resources provided by NVIDIA, the Ohio Supercomputer Center (OSC)
and through ACCESS (allocation CIS250105) and the National
Artificial Intelligence Research Resource (NAIRR) Pilot
(allocation 240239).

%% file: supp/0_overview.tex
\section{Supplementary Overview}
\label{sec:supp_overview}

This supplement provides implementation details for the iterative
triage procedure, calibration, acquisition strategies, dataset
construction, and baselines. We also include additional experimental
results, sensitivity analyses, runtime measurements, and qualitative
examples. Unless otherwise stated, experiments use the same frozen
representations, proposal pools, calibration protocol, review budgets,
and five random seeds as the main paper.

%% file: supp/1_method_details.tex
\section{Complete Method and Algorithm Details}
\label{sec:supp_method}

This section specifies the complete iterative acquisition-and-triage
procedure. The detector and visual encoders remain frozen throughout the
process. Expert-reviewed proposals are reused directly as non-parametric
prototypes, while automatically handled proposals do not become new
supervision.

\subsection{Iterative Acquisition and Triage}
\label{sec:supp_iterative}

Let $\mathcal{U}$ denote a fixed pool of object proposals and
$y_x\in\{0,1,\ldots,C\}$ the label of proposal $x$, where $0$ denotes
background and $1,\ldots,C$ denote foreground classes. For every proposal,
we precompute two $\ell_2$-normalized frozen representations,
$\mathbf{z}^{\mathrm{fg}}_x$ and $\mathbf{z}^{\mathrm{bg}}_x$.
Under the canonical \emph{dual-encoder configuration}, these correspond to
BioCLIP2~\cite{gu2025bioclip} and DINOv3~\cite{simeoni2025dinov3},
respectively. Acquisition is performed in the BioCLIP2 foreground
representation space.

Let $\mathcal{P}_t$ denote the expert-reviewed prototype bank and
$\mathcal{R}_t$ the unresolved pool at round $t$. The prototype bank is
initially empty. At each round, an acquisition strategy selects a batch
$\mathcal{B}_t\subseteq\mathcal{R}_t$ for expert review. The reviewed
proposals and their labels are added to the prototype bank, after which all
remaining proposals are re-triaged as \emph{Resolve}, \emph{Clear}, or
\emph{Defer}. Reviewed, Resolved, and Cleared proposals leave the unresolved
pool; only Deferred proposals remain eligible for subsequent acquisition.
Automatically Resolved or Cleared proposals are never added to the prototype
bank.

Algorithm~\ref{alg:active_triage} summarizes the complete procedure.

\begin{algorithm}[t]
\caption{Training-Free Active Prototype Triage}
\label{alg:active_triage}
\begin{algorithmic}[1]
\Require Proposal pool $\mathcal{U}$; frozen representations
$Z^{\mathrm{fg}},Z^{\mathrm{bg}}$; labeled calibration bank $\mathcal{C}$;
expert oracle $\mathcal{O}$; acquisition rule $A$; batch size $B$;
maximum rounds $T$
\State Calibrate $k^\star$ on $\mathcal{C}$
\State $\tau_{\mathrm{fg}},\tau_{\mathrm{bg}}
    \gets \lfloor 2k^\star/3\rfloor+1$
\State $\mathcal{P}\gets\emptyset$
\State $\mathcal{R}\gets\mathcal{U}\setminus\mathcal{C}$
\For{$t=1,\ldots,T$}
    \If{$\mathcal{R}=\emptyset$}
        \State \textbf{break}
    \EndIf
    \State $\mathcal{B}_t\gets A(\mathcal{R},\mathcal{P},B)$
    \State Obtain expert labels for $\mathcal{B}_t$
    \State $\mathcal{P}\gets\mathcal{P}\cup\mathcal{B}_t$
    \State Recompute class-specific floors $\{\rho_c\}_{c=1}^{C}$
    \State $\mathcal{D}_t\gets\emptyset$
    \ForAll{$x\in\mathcal{R}\setminus\mathcal{B}_t$}
        \If{\Call{ForegroundRule}{$x,\mathcal{P}$}}
            \State \emph{Resolve} $x$ to class $\hat c(x)$
        \ElsIf{\Call{BackgroundRule}{$x,\mathcal{P}$}}
            \State \emph{Clear} $x$ as background
        \Else
            \State $\mathcal{D}_t\gets\mathcal{D}_t\cup\{x\}$
        \EndIf
    \EndFor
    \State $\mathcal{R}\gets\mathcal{D}_t$
\EndFor
\end{algorithmic}
\end{algorithm}

In our retrospective experiments, the expert oracle is simulated using the
available reference annotations. The calibration bank is separate from both
the active-review pool and the evaluation pool.

\subsection{Resolve, Clear, and Defer}
\label{sec:supp_triage_rule}

For proposal $x$, let $\mathcal{N}^{\mathrm{fg}}_k(x)$ and
$\mathcal{N}^{\mathrm{bg}}_k(x)$ denote its nearest verified prototypes in
the foreground and background representation spaces. Neighbors are ranked by
cosine similarity. If fewer than the calibrated $k$ prototypes are available,
all available prototypes are used, but the decision thresholds remain defined
by the calibrated value of $k$:
\begin{equation}
    \tau_{\mathrm{fg}}
    =
    \tau_{\mathrm{bg}}
    =
    \left\lfloor\frac{2k}{3}\right\rfloor+1.
    \label{eq:supp_vote_threshold}
\end{equation}

\paragraph{Foreground evidence.}
Each foreground-labeled prototype in
$\mathcal{N}^{\mathrm{fg}}_k(x)$ votes for its verified class. For class $c$,
\[
    v_c(x)
    =
    \sum_{p\in\mathcal{N}^{\mathrm{fg}}_k(x)}
    \mathbf{1}[y_p=c].
\]
Let $\hat c(x)$ denote the unique foreground class receiving the largest
positive vote count, when such a class exists, and let
$v_{\mathrm{fg}}(x)=v_{\hat c(x)}(x)$. If no foreground-labeled prototype
appears among the neighbors, or if the largest foreground vote is tied, the
Resolve rule is not satisfied.

\paragraph{Class-specific similarity floor.}
Let
\[
    \mathcal{P}_c=\{p\in\mathcal{P}:y_p=c\}
\]
be the reviewed prototypes for foreground class $c$. For each
$p\in\mathcal{P}_c$, define its nearest leave-one-out same-class similarity
as
\[
    \eta(p)
    =
    \max_{p'\in\mathcal{P}_c\setminus\{p\}}
    \cos\!\left(
        \mathbf{z}^{\mathrm{fg}}_p,
        \mathbf{z}^{\mathrm{fg}}_{p'}
    \right).
\]

For classes with at least two reviewed prototypes,
\begin{equation}
    \rho_c
    =
    \operatorname{Quantile}_{q}
    \left(
        \{\eta(p):p\in\mathcal{P}_c\}
    \right).
    \label{eq:supp_similarity_floor}
\end{equation}
The canonical configuration uses $q=0$, so $\rho_c$ is the minimum
leave-one-out nearest same-class similarity. Classes with fewer than two
reviewed prototypes are not eligible for automatic Resolve.

For query $x$, its strongest similarity to class $c$ is
\[
    s_{\mathrm{fg}}(x,c)
    =
    \max_{p\in\mathcal{P}_c}
    \cos\!\left(
        \mathbf{z}^{\mathrm{fg}}_x,
        \mathbf{z}^{\mathrm{fg}}_p
    \right).
\]

If no such unique winner exists, $F(x)=\mathrm{false}$. Otherwise,
\begin{equation}
\begin{aligned}
    F(x)
    ={}&
    \left[v_{\mathrm{fg}}(x)\ge\tau_{\mathrm{fg}}\right]
    \wedge
    \left[|\mathcal{P}_{\hat c(x)}|\ge2\right]
    \\
    &\wedge
    \left[
        s_{\mathrm{fg}}(x,\hat c(x))
        \ge\rho_{\hat c(x)}
    \right].
\end{aligned}
\label{eq:supp_fg_rule}
\end{equation}

\paragraph{Background evidence.}
Background support is computed independently in the background
representation space:
\begin{equation}
    v_{\mathrm{bg}}(x)
    =
    \sum_{p\in\mathcal{N}^{\mathrm{bg}}_k(x)}
    \mathbf{1}[y_p=0].
    \label{eq:supp_bg_votes}
\end{equation}

Foreground resolution is evaluated before background clearing:
\begin{equation}
\operatorname{Triage}(x)=
\begin{cases}
    \operatorname{Resolve}(\hat c(x)),
        & F(x),\\[1mm]
    \operatorname{Clear},
        & \neg F(x)\wedge
          v_{\mathrm{bg}}(x)\ge\tau_{\mathrm{bg}},\\[1mm]
    \operatorname{Defer},
        & \text{otherwise}.
\end{cases}
\label{eq:supp_triage}
\end{equation}

\subsection{Calibration}
\label{sec:supp_calibration}

The neighborhood size $k$ is selected once before active review using a
seed-specific labeled calibration bank of 100 proposals. Calibration
proposals are excluded from acquisition and evaluation and are not added to
the active prototype bank.

We consider
\[
    k\in\{5,7,9,11\},
\]
with the corresponding strict two-thirds vote thresholds given by
Eq.~\ref{eq:supp_vote_threshold}. Each candidate is evaluated by
leave-one-out triage on the calibration bank.

Candidate settings are ordered lexicographically by foreground-to-background
error count, foreground-to-background error rate, higher Correct Auto, lower
Deferred rate, and smaller $k$. The implementation additionally includes vote
thresholds as deterministic tie-break fields; under the canonical protocol,
these are fixed functions of $k$.

The selected $k$ remains fixed throughout the active-review run, while the
class-specific floors $\rho_c$ evolve with the prototype bank.

\subsection{Active Prototype Acquisition}
\label{sec:supp_acquisition}

All acquisition strategies operate on frozen BioCLIP2 proposal embeddings
\cite{gu2025bioclip}. Let $\mathcal{R}_t$ denote the current unresolved pool
and $\mathcal{S}$ the proposals already selected within the current
acquisition batch.

\subsubsection{Random and Batchwise $k$-center}

Random acquisition samples unresolved proposals uniformly without
replacement.

Batchwise $k$-center follows the Core-Set farthest-first selection principle
\cite{sener2018activelearningconvolutionalneural}. It applies farthest-first
traversal within each acquisition batch. The first proposal is the unresolved
proposal nearest to the unresolved-pool centroid. Subsequent proposals
maximize
\begin{equation}
    d(i,\mathcal{S})
    =
    1-
    \max_{j\in\mathcal{S}}
    \cos(\mathbf{z}_i,\mathbf{z}_j).
    \label{eq:supp_kcenter}
\end{equation}
This baseline is not warm-started from previously reviewed prototypes.

\subsubsection{Density-Aware k-center}

Motivated by density-aware coverage and active-selection approaches
\cite{climaco2026densityawarefarthestpointsampling,yan2026beyond,DACS_2022}, local density is estimated
from the mean cosine similarity to neighboring proposals in BioCLIP2 space,
using up to 30 nearest neighbors. Density values are converted to percentile
ranks $r_i\in[0,1]$, with larger values corresponding to denser regions. For
DW and DM, density is computed once over the initial acquisition pool and
reused throughout the active trajectory.

Density-aware acquisition scores proposal $i$ as
\begin{equation}
    a_i=d(i,\mathcal{S})\,w(r_i),
    \label{eq:supp_weighted_kcenter}
\end{equation}
with
\begin{equation}
w(r)=
\begin{cases}
    r,       & \text{Density-weighted (DW)},\\
    r(1-r),  & \text{Density-moderated (DM)}.
\end{cases}
\label{eq:supp_density_weights}
\end{equation}
Weighted selection initializes each batch using the proposal with the largest
density weight and subsequently maximizes
Eq.~\ref{eq:supp_weighted_kcenter}. The DM density weight therefore favors
intermediate-density proposals while retaining the farthest-first coverage
term.

\subsubsection{TypiClust}

TypiClust~\cite{hacohen2022activelearningbudgetopposite} clusters previously
reviewed and currently unresolved proposals jointly in BioCLIP2 space.
Clusters containing fewer reviewed examples are prioritized, with larger
clusters preferred when review counts are equal. Within each selected cluster,
unresolved proposals are ranked by typicality:
\begin{equation}
    \operatorname{typ}(x)
    =
    \left(
        \frac{1}{|\mathcal{N}_{K'}(x)|}
        \sum_{x'\in\mathcal{N}_{K'}(x)}
        \|\mathbf{z}_x-\mathbf{z}_{x'}\|_2
        +10^{-5}
    \right)^{-1},
    \label{eq:supp_typiclust}
\end{equation}
where $\mathcal{N}_{K'}(x)$ contains within-cluster nearest neighbors.
The neighborhood size is capped by the available same-cluster points, with a
canonical maximum of $K'=20$.

\subsubsection{Sampled Facility Location}

Facility Location follows the submodular representativeness objective used
for subset selection and active learning
\cite{Submodularity_facility_2015}. For each acquisition batch, Facility
Location considers at most 1024 sampled unresolved proposals as candidate
facilities. If fewer than 1024 proposals remain, the complete unresolved pool
is used.

Let $\mathcal{S}$ denote the facilities already selected in the current batch
and
\[
    m_i
    =
    \max_{j\in\mathcal{S}}
    \cos(\mathbf{z}_i,\mathbf{z}_j)
\]
the current similarity coverage of proposal $i$, with $m_i=-1$ before the
first facility is selected. Candidate $f$ is selected greedily according to
its facility-location marginal gain
\begin{equation}
    \Delta(f)
    =
    \sum_{i\in\mathcal{R}_t}
    \left[
        \max\!\left(
            \cos(\mathbf{z}_i,\mathbf{z}_f),
            m_i
        \right)
        -m_i
    \right].
    \label{eq:supp_facility}
\end{equation}
At each step, the candidate with the largest marginal gain is selected and
the coverage values are updated.

\subsubsection{DACS-TF}

DACS-TF adapts Density-Aware Core-Set
\cite{DACS_2022} to the frozen-feature setting. Unlike DW and DM, which reuse
density estimates computed on the initial acquisition pool, DACS-TF recomputes
local density over the current unresolved pool $\mathcal{R}_t$ at every
acquisition round.

Local density is computed from mean nearest-neighbor cosine similarity in
BioCLIP2 space using up to 30 nearest neighbors. The resulting density values
are partitioned into up to four one-dimensional density strata, ordered from
sparse to dense. When fewer than four nonempty strata can be formed, only the
available strata are used.

Let $m_g$ denote the number of currently unresolved proposals in stratum $g$,
and let $m=|\mathcal{R}_t|$. The target fraction of the acquisition budget
assigned to stratum $g$ is
\begin{equation}
    \pi_g
    =
    \frac{
        \exp\!\left((1-m_g/m)/\tau\right)
    }{
        \sum_h
        \exp\!\left((1-m_h/m)/\tau\right)
    },
    \qquad \tau=0.25.
    \label{eq:supp_dacs_allocation}
\end{equation}
Thus, smaller density strata receive larger relative allocation weights.

The fractional allocations $\pi_g B$ are converted to integer per-stratum
budgets using largest-remainder rounding, subject to the number of unresolved
proposals available in each stratum. Per-stratum allocations are therefore
capped by stratum capacity. Strata are processed from sparse to dense.

Within each stratum, DACS-TF performs farthest-first $k$-center selection in
BioCLIP2 space. Previously reviewed proposals are used to warm-start the
coverage set. If no previously reviewed proposal is available, the first
unresolved proposal in the stratum is chosen as the one nearest to that
stratum's centroid; subsequent proposals maximize distance to the current
coverage set.

If the stratum-specific selections do not fill the requested acquisition batch,
the remaining slots are filled by batchwise $k$-center over the still-unselected
proposals in the full unresolved pool.

%% file: supp/2_evaluation_protocol.tex
\section{Evaluation and Reproducibility Details}
\label{sec:supp_evaluation}

\subsection{Datasets and Proposal-Pool Construction}
\label{sec:supp_datasets}

We evaluate on two camera-trap and two aerial wildlife proposal pools.
Table~\ref{tab:supp_dataset_summary} reports the final pool statistics
after proposal generation, ground-truth assignment, and dataset-specific
filtering.

\begin{table*}[t]
\centering
\caption{
\textbf{Evaluation proposal pools.}
FG and BG denote foreground and background proposals after assignment
to reference annotations. FG prevalence is measured before removal
of the seed-specific calibration bank.
}
\label{tab:supp_dataset_summary}
\small
\begin{tabular}{lrrrrrr}
\toprule
Dataset & Images & Proposals & FG classes & FG proposals & BG proposals & FG (\%) \\
\midrule
ENA24       & 8,650  & 11,165 & 21 & 9,641  & 1,524 & 86.4 \\
LoTE-Animal & 19,787 & 39,690 & 11 & 31,607 & 8,083 & 79.6 \\
MMLA-Mpala  & 1,540  & 29,150 & 2  & 19,219 & 9,931 & 65.9 \\
WAUD        & 651    & 19,840 & 3  & 16,482 & 3,358 & 83.1 \\
\bottomrule
\end{tabular}
\end{table*}

\paragraph{ENA24.}
ENA24~\cite{yousif2019dynamic} contains 21 foreground classes,
providing a multiclass camera-trap setting.

\paragraph{LoTE-Animal.}
LoTE-Animal~\cite{Liu_2023_ICCV} provides an imbalanced multiclass
camera-trap setting. Because evaluation is transductive over a fixed
proposal pool, we combine the training and test splits. Labels remain
hidden except during calibration and simulated expert review.

\paragraph{MMLA-Mpala.}
We use an aerial subset of MMLA-Mpala
\cite{kline2025mmla_dataset,kline2025mmla}, derived from the KABR Raw Videos
collection~\cite{KABR_Raw_Videos}. We use Session~5, which
contains four drone videos depicting giraffes, Grevy's zebras, and Plains
zebras. Grevy's zebra and Plains zebra annotations are merged into a
single \emph{zebra} foreground class. To reduce temporal redundancy,
we retain every 30th frame.

For the selected frames, we manually refine the released bounding boxes
to more tightly fit the visible animals. We also observed occasional
zebras and giraffes far from the primary herd that occupy only a few
pixels and are absent from the released annotations. These instances
are difficult to identify even under manual inspection and can cause
valid detector proposals to be counted as background.

\paragraph{WAUD.}
Wilds Aerial Ungulate Detection (WAUD) combines onager imagery and
annotations from MMLA-Wilds
\cite{kline2025mmla,mmla_wilds,kline2025wildwing}
with SmartWilds aerial imagery
\cite{kline2025smartwildsmultimodalwildlifemonitoring,thewilds_drones_2025}.

For the MMLA-Wilds component, we use Session~2, comprising seven source
videos, and retain every 60th frame to reduce temporal redundancy.
For SmartWilds, we retain every 120th frame because camera motion is
comparatively limited while animals continue to move within the field
of view.

All selected SmartWilds frames were manually annotated for this study
with tight bounding boxes around visible animals. The SmartWilds
component contains P\`ere David's deer and Przewalski's horse, while
MMLA-Wilds contributes onager imagery. WAUD contains three
visually related ungulate classes with substantial foreground imbalance.

\subsection{Proposal Generation and Assignment}
\label{sec:supp_proposals}

Object proposals are generated with SAM3~\cite{carion2026sam3segmentconcepts}
using the text prompt \emph{animal}. We retain detections with confidence
at least $0.4$ before non-maximum suppression and apply NMS at IoU $0.5$.
Camera-trap datasets are processed without tiled inference.

For aerial imagery containing small animals, we use SAHI-style tiled
inference~\cite{sahi_2022} with an overlap ratio of $0.25$.
MMLA-Mpala and the MMLA-Wilds onager imagery use
$2016\times2016$-pixel slices, while SmartWilds imagery containing
P\`ere David's deer uses $1008\times1008$-pixel slices.
SmartWilds imagery containing Przewalski's horse is processed
without tiled inference because the animals occupy sufficiently
large image regions.

Each retained proposal is assigned to the highest-overlap reference
ground-truth box when intersection-over-union is at least $0.25$.
Otherwise, it is assigned the background label. For WAUD,
ground-truth boxes and proposals whose minimum box side is below
16 pixels are excluded before evaluation. No corresponding
minimum-side filter is applied to the other three pools.

\paragraph{Proposal-generator comparison.}
For comparison with SAM3, we use MegaDetector
v6~\cite{beery2019efficientpipelinecameratrap} with the
\texttt{MDV6-yolov10-e} animal detector and a confidence
threshold of $0.2$.

\subsection{Foreground Class Distributions}
\label{sec:supp_class_distribution}

Table~\ref{tab:supp_class_distribution} reports the foreground
proposal counts and class proportions for all four datasets.
LoTE-Animal dataset labels are retained, with common names
provided in parentheses.

\begin{table*}[t]
\centering
\caption{
\textbf{Foreground class distribution.}
Counts and percentages are computed over foreground proposals
in each final proposal pool. LoTE-Animal common names are
included in parentheses.
}
\label{tab:supp_class_distribution}
\small
\begin{tabular}{lp{0.56\textwidth}rr}
\toprule
Dataset & Class / species & Count & FG (\%) \\
\midrule
ENA24 & horse                         & 62    & 0.64 \\
ENA24 & woodchuck                     & 206   & 2.14 \\
ENA24 & bird                          & 225   & 2.33 \\
ENA24 & northern raccoon              & 292   & 3.03 \\
ENA24 & striped skunk                 & 303   & 3.14 \\
ENA24 & eastern chipmunk              & 309   & 3.21 \\
ENA24 & eastern gray squirrel         & 320   & 3.32 \\
ENA24 & bobcat                        & 333   & 3.45 \\
ENA24 & eastern cottontail            & 342   & 3.55 \\
ENA24 & coyote                        & 344   & 3.57 \\
ENA24 & eastern fox squirrel          & 346   & 3.59 \\
ENA24 & white-tailed deer             & 382   & 3.96 \\
ENA24 & red fox                       & 413   & 4.28 \\
ENA24 & wild turkey                   & 428   & 4.44 \\
ENA24 & gray fox                      & 441   & 4.57 \\
ENA24 & domestic cat                  & 481   & 4.99 \\
ENA24 & dog                           & 720   & 7.47 \\
ENA24 & Virginia opossum              & 726   & 7.53 \\
ENA24 & chicken                       & 734   & 7.61 \\
ENA24 & American black bear           & 960   & 9.96 \\
ENA24 & American crow                 & 1,274 & 13.21 \\
\midrule
LoTE-Animal & \textit{Martes flavigula} (yellow-throated marten)
    & 316 & 1.00 \\
LoTE-Animal & \textit{Hystrix brachyura} (Malayan porcupine)
    & 685 & 2.17 \\
LoTE-Animal & \textit{Capricornis milneedwardsii} (Chinese serow)
    & 903 & 2.86 \\
LoTE-Animal & RedPandas (red panda)
    & 954 & 3.02 \\
LoTE-Animal & \textit{Sus scrofa} (wild boar)
    & 1,528 & 4.83 \\
LoTE-Animal & Pandas (giant panda)
    & 1,744 & 5.52 \\
LoTE-Animal & \textit{Rhinopithecus roxellana} (golden snub-nosed monkey)
    & 1,958 & 6.19 \\
LoTE-Animal & \textit{Elaphodus cephalophus} (tufted deer)
    & 2,177 & 6.89 \\
LoTE-Animal & \textit{Rusa unicolor} (sambar deer)
    & 3,175 & 10.05 \\
LoTE-Animal & \textit{Macaca thibetana} (Tibetan macaque)
    & 5,821 & 18.42 \\
LoTE-Animal & \textit{Pseudois nayaur} (bharal / blue sheep)
    & 12,346 & 39.06 \\
\midrule
MMLA-Mpala & giraffe
    & 3,314 & 17.24 \\
MMLA-Mpala & zebra
    & 15,905 & 82.76 \\
\midrule
WAUD & Przewalski's horse
    & 1,205 & 7.31 \\
WAUD & onager
    & 1,512 & 9.17 \\
WAUD & P\`ere David's deer
    & 13,765 & 83.52 \\
\bottomrule
\end{tabular}
\end{table*}

\subsection{Frozen Proposal Representations}
\label{sec:supp_features}

Proposal representations are extracted once before active review and cached
for all subsequent experiments. No detector or encoder forward passes are
required during the active loop.

Proposal crops correspond to the clipped proposal bounding box without
additional spatial padding. BioCLIP2~\cite{gu2025bioclip} produces
768-dimensional representations, while
DINOv3-Huge+~\cite{simeoni2025dinov3} produces 1280-dimensional
representations. Features used for cosine similarity are $\ell_2$ normalized.

For DINOv3-Huge+, proposal crops are resized and normalized before
feature extraction. Patch features are normalized, spatially
mean-pooled, and the resulting proposal representation is normalized
again. Neither encoder is updated during active review, and no
downstream classifier is fitted by Prototype Triage.

\subsection{Per-Seed Calibrated Neighborhood Sizes}
\label{sec:supp_calibrated_k}

Table~\ref{tab:supp_calibrated_k} reports the selected neighborhood
sizes for each dataset, seed, and foreground/background encoder
configuration. Each configuration is calibrated independently
using the protocol in Sec.~\ref{sec:supp_calibration}.
The corresponding vote thresholds follow
Eq.~\ref{eq:supp_vote_threshold}.

\begin{table*}[t]
\centering
\caption{
\textbf{Per-seed calibrated neighborhood size $k$.}
Each entry is selected independently from $\{5,7,9,11\}$ using the
100-proposal calibration bank for the corresponding dataset, seed,
and encoder configuration.
$\star$ marks the dual-encoder configuration used in the remaining experiments.
}
\label{tab:supp_calibrated_k}
\small
\begin{tabular}{lllccccc}
\toprule
Dataset & FG encoder & BG encoder & Seed 3 & Seed 7 & Seed 11 & Seed 13 & Seed 17 \\
\midrule
ENA24
    & DINOv3           & DINOv3   & 7 & 5 & 5 & 5 & 5 \\
    & BioCLIP2         & BioCLIP2 & 9 & 5 & 9 & 5 & 9 \\
    & DINOv3           & BioCLIP2 & 9 & 5 & 9 & 5 & 9 \\
\cmidrule(lr){2-8}
    & $\star$ BioCLIP2 & DINOv3   & 7 & 5 & 5 & 5 & 5 \\
\midrule
LoTE-Animal
    & DINOv3           & DINOv3   & 5 & 11 & 7 & 7 & 7 \\
    & BioCLIP2         & BioCLIP2 & 9 & 9 & 7 & 9 & 11 \\
    & DINOv3           & BioCLIP2 & 7 & 9 & 7 & 7 & 11 \\
\cmidrule(lr){2-8}
    & $\star$ BioCLIP2 & DINOv3   & 5 & 9 & 7 & 7 & 7 \\
\midrule
WAUD
    & DINOv3           & DINOv3   & 5 & 5 & 7 & 5 & 7 \\
    & BioCLIP2         & BioCLIP2 & 7 & 11 & 5 & 9 & 5 \\
    & DINOv3           & BioCLIP2 & 7 & 5 & 5 & 7 & 5 \\
\cmidrule(lr){2-8}
    & $\star$ BioCLIP2 & DINOv3   & 7 & 5 & 7 & 5 & 5 \\
\midrule
MMLA-Mpala
    & DINOv3           & DINOv3   & 5 & 7 & 7 & 5 & 7 \\
    & BioCLIP2         & BioCLIP2 & 5 & 5 & 5 & 11 & 5 \\
    & DINOv3           & BioCLIP2 & 11 & 5 & 7 & 5 & 5 \\
\cmidrule(lr){2-8}
    & $\star$ BioCLIP2 & DINOv3   & 5 & 7 & 7 & 7 & 7 \\
\bottomrule
\end{tabular}
\end{table*}

\subsection{Evaluation Metrics}
\label{sec:supp_metrics}

We report three primary metrics, two automatic-error diagnostics,
and additional supplementary measures.
All metrics are computed on the calibration-excluded evaluation pool.
Let $N$ be the number of evaluation proposals, $N_{\mathrm{FG}}$
the number of true-foreground proposals, $H$ the number reviewed,
and $D$ the number remaining Deferred.

\paragraph{Review Burden.}
Review Burden is the fraction of evaluation proposals that have been
reviewed or remain Deferred:
\begin{equation}
    \mathrm{ReviewBurden}
    =
    \frac{H+D}{N}.
    \label{eq:supp_review_burden}
\end{equation}
It measures proposal-level annotation demand rather than annotation time.

\paragraph{Automatic FG Recall.}
Automatic FG Recall measures the fraction of true-foreground proposals
correctly Resolved without expert review:
\begin{equation}
    \mathrm{AutoFGRecall}
    =
    \frac{
        N_{\mathrm{correct\ automatic\ FG\ Resolve}}
    }{
        N_{\mathrm{FG}}
    }.
    \label{eq:supp_auto_fg_recall}
\end{equation}
Reviewed, Deferred, incorrectly Cleared, and incorrectly classified
foreground proposals remain in the denominator.

\paragraph{Conditional FG Macro-F1.}
Let $\mathcal{I}_{\mathrm{FG}}$ denote true-foreground proposals
automatically Resolved to a foreground class. The macro-F1 label set
$\mathcal{C}_{\mathrm{eval}}$ contains every non-background class ID
present in the full evaluation ground-truth or prediction vector.

For each class $c$, precision and recall counts are computed only
over $\mathcal{I}_{\mathrm{FG}}$. Conditional FG Macro-F1 is
\begin{equation}
    \mathrm{CondFGMacroF1}
    =
    \frac{1}{|\mathcal{C}_{\mathrm{eval}}|}
    \sum_{c\in\mathcal{C}_{\mathrm{eval}}}
    \frac{2TP_c}{2TP_c+FP_c+FN_c}.
    \label{eq:supp_cond_fg_f1}
\end{equation}
A class with a zero denominator contributes zero
(\texttt{zero\_division=0}). If $\mathcal{I}_{\mathrm{FG}}$
is empty, the metric is undefined.

\paragraph{False Clear Rate.}
False Clear Rate is the fraction of true-foreground proposals
incorrectly Cleared as background:
\begin{equation}
    \mathrm{FalseClearRate}
    =
    \frac{
        N_{\mathrm{true\ FG\ Cleared\ as\ BG}}
    }{
        N_{\mathrm{FG}}
    }.
    \label{eq:supp_false_clear}
\end{equation}

\paragraph{Automatic Error Rate.}
Automatic Error Rate is the fraction of automatic Resolve and Clear
decisions that are incorrect:
\begin{equation}
    \mathrm{AutomaticErrorRate}
    =
    \frac{
        N_{\mathrm{incorrect\ automatic}}
    }{
        N_{\mathrm{automatic\ Resolve}}
        +N_{\mathrm{automatic\ Clear}}
    }.
    \label{eq:supp_auto_error}
\end{equation}
Errors include wrong foreground assignments, foreground proposals
Cleared as background, and background proposals Resolved as foreground.
We use this metric for the explicit-deferral comparison.

\paragraph{Correct Auto.}
Correct Auto measures correct automatic decisions as a fraction
of the full evaluation pool:
\begin{equation}
    \mathrm{CorrectAuto}
    =
    \frac{
        N_{\mathrm{correct\ Resolve}}
        +N_{\mathrm{correct\ Clear}}
    }{
        N
    }.
    \label{eq:supp_correct_auto}
\end{equation}
Unlike Automatic FG Recall, it includes correct background clearing.

\paragraph{Wrong-FG.}
Wrong-FG measures incorrect foreground-class assignments as a fraction
of true-foreground proposals:
\begin{equation}
    \mathrm{WrongFG}
    =
    \frac{
        N_{\mathrm{true\ FG\ Resolved\ to\ wrong\ FG}}
    }{
        N_{\mathrm{FG}}
    }.
    \label{eq:supp_wrong_fg}
\end{equation}
It distinguishes incorrect species assignments from False Clear errors.

\paragraph{Deferred (\%).}
For acquisition trajectories, Deferred (\%) is the percentage of
evaluation proposals remaining Deferred after a given review budget:
\begin{equation}
    \mathrm{Deferred}(t)
    =
    \frac{D_t}{N}\times100.
    \label{eq:supp_deferred}
\end{equation}
Unlike Review Burden, it excludes proposals already reviewed.

%% file: supp/3_baseline_details.tex
\section{Baseline and Ablation Details}
\label{sec:supp_baselines}

We describe the Symmetric $k$-NN and Uncertainty + logistic
regression (LR) baselines, followed by the forced-decision
rule used in the Defer ablation. Both baselines use frozen
BioCLIP2 representations and treat foreground classes
and background as multiclass labels.

\subsection{Symmetric \texorpdfstring{$k$}{k}-NN Implementation}
\label{sec:supp_selective_knn}

Symmetric $k$-NN reuses the same DM-acquired reviewed
proposals as dual-encoder Prototype Triage for each dataset
and seed. At round $t$, $\mathcal{L}_t$ contains all proposals
reviewed through that round.

We consider $k\in\{5,7,9,11\}$ and retrieve the
$k_{\mathrm{eff}}=\min(k,|\mathcal{L}_t|)$ nearest reviewed
prototypes by cosine similarity in BioCLIP2 space.
Foreground classes and background participate in the
same multiclass vote. Let $n_c(x)$ denote the number of
neighbors voting for class $c$. The winning-class confidence is
\begin{equation}
    \operatorname{conf}_{\mathrm{kNN}}(x)
    =
    \frac{\max_c n_c(x)}{k_{\mathrm{eff}}}.
    \label{eq:supp_knn_confidence}
\end{equation}

Vote-count ties are resolved by the summed cosine
similarity of tied-class neighbors, then by their maximum
cosine similarity, with any remaining exact ties resolved
deterministically. Proposals with confidence at least
$\gamma$ receive the predicted class label; others are Deferred.

\subsection{Selective-Risk Calibration}
\label{sec:supp_selective_calibration}

Symmetric $k$-NN jointly calibrates $k$ and the confidence
threshold $\gamma$ using the fixed, seed-specific
100-proposal calibration bank. These examples are excluded
from active review and evaluation.

For automatically predicted calibration proposals
$\mathcal{A}_{\mathrm{cal}}$, empirical selective error is
\begin{equation}
    R_{\mathrm{sel}}
    =
    \frac{
        \sum_{x\in\mathcal{A}_{\mathrm{cal}}}
        \mathbf{1}[\hat y(x)\neq y_x]
    }{
        |\mathcal{A}_{\mathrm{cal}}|
    }.
    \label{eq:supp_selective_risk}
\end{equation}

Settings with zero automatic coverage are excluded from
the primary feasible set. Candidate thresholds include
distinct confidence values observed on the calibration
bank and boundary thresholds for zero and full coverage.

We target $R_{\mathrm{sel}}\le0.05$. Among feasible
nonzero-coverage settings, calibration maximizes automatic
coverage, minimizes foreground-to-background errors,
and then prefers smaller $k$ and stricter confidence
thresholds. If no setting satisfies the error target,
calibration minimizes selective error first, followed by
coverage, foreground safety, and the same tie-breakers.

Calibration is repeated at each round using the current
reviewed support and the fixed calibration bank. Evaluation
labels are never used for parameter selection.

\paragraph{Paired-seed consistency.}
Symmetric $k$-NN and Prototype Triage use identical
DM-acquired reviewed support for each dataset and seed.
Across the 20 dataset--seed pairs, Prototype Triage has
lower Review Burden and higher Automatic FG Recall in
19/20 pairs. LoTE-Animal seed~11 is the only exception
for both metrics.

\subsection{Logistic-Regression Implementation}
\label{sec:supp_lr}

The initial LR review batch matches the seed-specific
initial batch of Random acquisition. Review remains random
until at least two classes are represented in the reviewed
set.

A multinomial logistic-regression model is then refitted
from scratch after each round using all reviewed examples.
We use $C=1.0$, \texttt{solver=lbfgs}, and
\texttt{max\_iter=5000}, without class weighting or an
additional probability-calibration stage.

The predicted label is the class with the highest posterior
probability, which also defines prediction confidence.
The confidence threshold is recalibrated each round using
the same $5\%$ selective-risk criterion in
Sec.~\ref{sec:supp_selective_calibration}. Calibration
examples are excluded from classifier training.

Automatic predictions are locked once made. Proposals
below the confidence threshold remain Deferred, and
those with the highest predictive entropy are selected
for the next review batch.

\subsection{Forced-Decision Rule}
\label{sec:supp_forced_decision}

The forced-decision ablation assigns labels only to
proposals remaining Deferred at the prototype-triage
endpoint. Previously reviewed labels, automatic decisions,
prototype banks, and frozen representations are unchanged.

For a Deferred proposal $x$, let $\hat c(x)$ denote
its unique foreground vote winner, when one exists.
Let $v_{\mathrm{fg}}(x)$ and $v_{\mathrm{bg}}(x)$ denote
the foreground-winner and background vote counts.
Define
\begin{align}
    G(x) ={}&
    [\,\hat c(x)\text{ is a unique positive winner}\,]
    \nonumber\\
    &\wedge
    [\,|\mathcal{P}_{\hat c(x)}|\ge2\,]
    \nonumber\\
    &\wedge
    [\,s_{\mathrm{fg}}(x,\hat c(x))
        \ge \rho_{\hat c(x)}\,]
    \nonumber\\
    &\wedge
    [\,v_{\mathrm{fg}}(x)\ge v_{\mathrm{bg}}(x)\,].
    \label{eq:supp_forced_predicate}
\end{align}

The forced prediction is
\begin{equation}
    \operatorname{Forced}(x)
    =
    \begin{cases}
        \hat c(x), & G(x),\\
        0, & \text{otherwise}.
    \end{cases}
    \label{eq:supp_forced_decision}
\end{equation}
Thus, Deferred proposals satisfying $G(x)$ are assigned
to their foreground winner; all others are assigned
to background.

%% file: supp/4_expanded_results.tex
\section{Expanded Experimental Results}
\label{sec:supp_expanded_results}

Unless otherwise stated, results are reported as mean $\pm$
standard deviation over five seeds, using the final available
round of each run. Runs terminate early when no Deferred
proposals remain.

\subsection{Foreground Classification Errors Across Representations}
\label{sec:supp_routing_results}

Table~\ref{tab:supp_routing_full} reports Wrong-FG for the four
foreground/background encoder configurations. Each configuration
is rerun end-to-end, with acquisition performed in its foreground
representation space and $k$ calibrated independently.

\begin{table*}[t]
\centering
\caption{
\textbf{Foreground misclassification by encoder configuration.}
Wrong-FG (\%, lower is better) measures the fraction of true-foreground
proposals automatically Resolved to an incorrect foreground class.
Column headings specify the FG/BG encoder pair. Values are five-seed
mean $\pm$ standard deviation.
$\star$ denotes the canonical dual-encoder configuration.
}
\label{tab:supp_routing_full}
\small
\setlength{\tabcolsep}{7pt}
\begin{tabular}{lcccc}
\toprule
Dataset &
\makecell{DINOv3 /\\DINOv3} &
\makecell{BioCLIP2 /\\BioCLIP2} &
\makecell{DINOv3 /\\BioCLIP2} &
\makecell{$\star$ BioCLIP2 /\\DINOv3} \\
\midrule
ENA24
    & $4.6{\pm}0.7$
    & $2.1{\pm}0.9$
    & $3.0{\pm}1.7$
    & $3.4{\pm}0.5$ \\
LoTE-Animal
    & $6.4{\pm}1.3$
    & $1.6{\pm}1.1$
    & $5.5{\pm}1.6$
    & $3.0{\pm}1.1$ \\
WAUD
    & $2.3{\pm}0.9$
    & $1.7{\pm}0.4$
    & $3.7{\pm}2.5$
    & $1.9{\pm}0.6$ \\
MMLA-Mpala
    & $0.1{\pm}0.1$
    & $0.1{\pm}0.0$
    & $0.1{\pm}0.0$
    & $0.2{\pm}0.1$ \\
\bottomrule
\end{tabular}
\end{table*}

BioCLIP2-only obtains the lowest or tied-lowest Wrong-FG on
all four pools. However, it automatically recovers fewer
true-foreground proposals than the dual-encoder configuration,
as shown in the main paper. Wrong-FG must therefore be considered
alongside Automatic FG Recall and False Clear Rate.

The differences are particularly pronounced on the multiclass
camera-trap pools. DINOv3-only has Wrong-FG of $4.6\%$ on
ENA24 and $6.4\%$ on LoTE-Animal, compared with $2.1\%$
and $1.6\%$ for BioCLIP2-only.

Because the foreground encoder also determines the acquisition
space, these comparisons measure the complete encoder
configuration rather than voting performance with a fixed
reviewed support.

\subsection{Additional Deferral Diagnostics}
\label{sec:supp_deferral_results}

Table~\ref{tab:supp_defer_full} reports additional error measures
for the explicit-Defer and forced-decision settings on the two
camera-trap pools. The forced rule is applied only to proposals
remaining Deferred at the endpoint; previously reviewed labels
and automatic decisions are unchanged.

\begin{table*}[t]
\centering
\caption{
\textbf{Effect of forcing Deferred proposals to a label.}
Correct Auto, Wrong-FG, and False Clear Rate are percentages.
Conditional FG Macro-F1 is reported on a 0--1 scale.
All values are five-seed mean $\pm$ standard deviation.
}
\label{tab:supp_defer_full}
\small
\setlength{\tabcolsep}{6pt}
\begin{tabular}{llcccc}
\toprule
Dataset & Setting &
\makecell{Correct\\Auto $\uparrow$} &
\makecell{Wrong-FG\\$\downarrow$} &
\makecell{False Clear\\Rate $\downarrow$} &
\makecell{Conditional\\FG Macro-F1 $\uparrow$} \\
\midrule
\multirow{2}{*}{ENA24}
& With Defer
& $74.0{\pm}0.6$
& $3.4{\pm}0.5$
& $2.3{\pm}0.7$
& $.943{\pm}.010$ \\
& Forced decision
& $79.4{\pm}0.2$
& $6.7{\pm}0.4$
& $7.3{\pm}0.3$
& $.903{\pm}.005$ \\
\midrule
\multirow{2}{*}{LoTE-Animal}
& With Defer
& $71.8{\pm}3.1$
& $3.0{\pm}1.1$
& $4.1{\pm}0.5$
& $.920{\pm}.024$ \\
& Forced decision
& $79.8{\pm}0.3$
& $7.2{\pm}0.3$
& $9.2{\pm}0.5$
& $.849{\pm}.008$ \\
\bottomrule
\end{tabular}
\end{table*}

Forced decision increases Correct Auto on both datasets,
but also increases incorrect foreground assignments and
erroneous background clearing. On ENA24, Wrong-FG rises
from $3.4\%$ to $6.7\%$ and False Clear Rate from
$2.3\%$ to $7.3\%$. On LoTE-Animal, the corresponding
increases are $3.0\%$ to $7.2\%$ and $4.1\%$ to $9.2\%$.
Conditional FG Macro-F1 also decreases on both pools.

The aerial pools have approximately zero residual Deferred
proposals at the endpoint. MMLA-Mpala has no remaining
Deferred proposals, while forced decision changes the WAUD
results negligibly after rounding. The effect of forced
decision is therefore concentrated on pools where substantial
unresolved workload remains.

\subsection{Endpoint Proposal-Outcome Decomposition}
\label{sec:supp_endpoint_outcome_decomposition}

Table~\ref{tab:supp_endpoint_outcome_decomposition} partitions
the evaluation pool into mutually exclusive endpoint outcomes
under canonical DM acquisition with the dual-encoder
configuration. The categories distinguish correct automatic
decisions, incorrect automatic decisions, Deferred proposals,
and reviewed proposals.

\begin{table*}[t]
\centering
\caption{
\textbf{Endpoint proposal-outcome decomposition.}
Five-seed mean $\pm$ standard deviation percentage of the
full evaluation pool in each mutually exclusive endpoint
category. Five-seed mean counts are shown in parentheses.
Reviewed denotes actively expert-reviewed proposals.
The separate 100-proposal calibration bank is excluded.
Totals may differ slightly from the sum of displayed
percentages because of rounding.
}
\label{tab:supp_endpoint_outcome_decomposition}
\footnotesize
\setlength{\tabcolsep}{3pt}
\resizebox{\textwidth}{!}{%
\begin{tabular}{lcccccccc}
\toprule
Dataset &
\makecell{Correct FG\\Resolve} &
\makecell{Wrong FG\\Resolve} &
\makecell{FG Cleared\\as BG} &
\makecell{Correct BG\\Clear} &
\makecell{BG Resolved\\as FG} &
Deferred &
Reviewed &
Total \\
\midrule
ENA24 &
$64.0{\pm}0.4$ (7078) &
$3.0{\pm}0.5$ (327) &
$2.0{\pm}0.6$ (216) &
$10.0{\pm}0.3$ (1110) &
$0.8{\pm}0.0$ (91) &
$13.0{\pm}1.2$ (1443) &
$7.2{\pm}0.0$ (800) &
$100.0$ \\
LoTE-Animal &
$59.2{\pm}2.5$ (23449) &
$2.4{\pm}0.9$ (961) &
$3.3{\pm}0.4$ (1307) &
$12.6{\pm}0.6$ (4996) &
$2.8{\pm}0.6$ (1116) &
$17.6{\pm}5.0$ (6961) &
$2.0{\pm}0.0$ (800) &
$100.0$ \\
WAUD &
$78.6{\pm}0.4$ (15508) &
$1.6{\pm}0.5$ (314) &
$0.9{\pm}0.3$ (171) &
$9.7{\pm}1.5$ (1911) &
$5.7{\pm}1.8$ (1117) &
$0.0{\pm}0.1$ (7) &
$3.6{\pm}0.5$ (711) &
$100.0$ \\
MMLA-Mpala &
$63.0{\pm}0.1$ (18307) &
$0.1{\pm}0.0$ (38) &
$1.7{\pm}0.1$ (488) &
$31.8{\pm}0.1$ (9226) &
$1.5{\pm}0.1$ (430) &
$0.0{\pm}0.0$ (0) &
$1.9{\pm}0.1$ (562) &
$100.0$ \\
\bottomrule
\end{tabular}%
}
\end{table*}

All percentages in this table use the full evaluation
pool as their denominator. In contrast, Wrong-FG and
False Clear Rate are normalized by the true-foreground
population. The corresponding values should not be
compared directly. ENA24 and LoTE-Animal retain Deferred
fractions of $13.0\%$ and $17.6\%$, respectively, after
800 active reviews. MMLA-Mpala has no remaining Deferred
proposals and a substantially larger correctly Cleared
background component ($31.8\%$ of its evaluation pool).

\begin{figure*}[t]
\centering
\includegraphics[width=\textwidth]{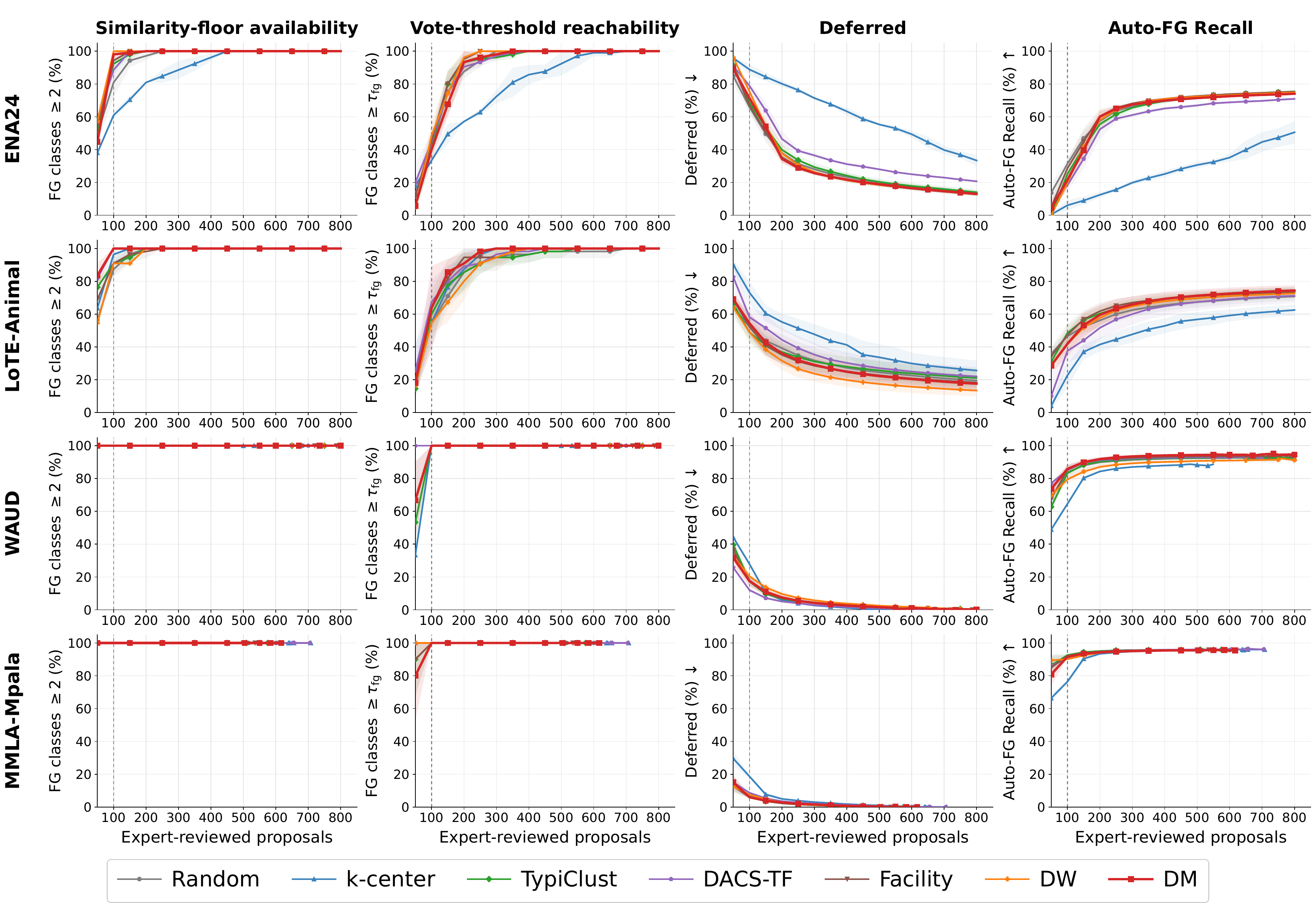}
\caption{
\textbf{Class support and triage outcomes across review budgets.}
Rows correspond to datasets. Columns show the percentage of foreground
classes with at least two reviewed prototypes (similarity-floor
availability), the percentage with at least $\tau_{\mathrm{fg}}$ reviewed
prototypes (vote-threshold reachability), the percentage of proposals
Deferred, and Automatic FG Recall. Curves show five-seed mean
trajectories. The first two measures indicate necessary support
conditions rather than successful automatic recognition.
}
\label{fig:supp_acquisition_support}
\end{figure*}

\subsection{Selective-Baseline Diagnostics}
\label{sec:supp_baseline_results}

Symmetric $k$-NN obtains higher Conditional FG Macro-F1
than Prototype Triage on several pools, but makes automatic
decisions on fewer proposals. On LoTE-Animal, its
Conditional FG Macro-F1 is $.963{\pm}.031$, compared with
$.920{\pm}.024$ for Prototype Triage, while Automatic
Coverage is $59.3{\pm}17.2\%$ versus $80.4{\pm}5.0\%$.
On WAUD, Automatic Coverage is $50.5{\pm}29.8\%$ for
Symmetric $k$-NN versus $96.4{\pm}0.6\%$ for Prototype
Triage.

These methods use the same DM-acquired reviewed support
within each dataset--seed pair. The uncertainty-sampling LR
baseline follows a separate acquisition trajectory and can
terminate after a different number of reviews.

\subsection{Class Support and Prototype Utility}
\label{sec:supp_prototype_utility}

A foreground class obtains a finite class-specific similarity floor after at
least two examples have been reviewed. Automatic \emph{Resolve} additionally
requires a unique foreground vote meeting the fixed threshold
$\tau_{\mathrm{fg}}$, so a class with fewer than $\tau_{\mathrm{fg}}$ reviewed
prototypes cannot satisfy the foreground voting condition.

Figure~\ref{fig:supp_acquisition_support} shows how class support and
downstream triage outcomes evolve with the review budget. On LoTE-Animal,
DM and DACS-TF obtain two-prototype support for every foreground class
within 100 active-review labels. Their subsequent Deferred and
Automatic FG Recall trajectories nevertheless differ. Similar
differences persist on ENA24 even after most strategies approach
complete vote-threshold reachability.

On WAUD, nearly all acquisition strategies reduce the remaining
Deferred pool to approximately zero, yet Automatic FG Recall ranges
from $88.4\%$ for batchwise $k$-center to $94.6\%$ for DM.
These results show that satisfying class-support requirements does
not guarantee effective foreground recognition. The examples
selected for review continue to matter after classes are represented.

%% file: supp/5_sensitivity.tex
\section{Sensitivity Analysis}
\label{sec:supp_sensitivity}

We examine sensitivity to the similarity-floor quantile $q$ and
neighborhood size $k$, which directly affect automatic triage
decisions. We also report shorter checks of calibration-bank size
and the density parameters used for acquisition. All experiments
use the frozen proposal representations and five-seed protocol
described in Sec.~\ref{sec:supp_evaluation}.

\subsection{Similarity-Floor Quantile}
\label{sec:supp_q_sensitivity}

The canonical class-specific similarity floor uses $q=0$, the
minimum leave-one-out same-class prototype similarity. We vary
\[
    q\in\{0,0.05,0.10,0.25\}.
\]
For each setting, class floors are updated as the reviewed
prototype bank grows. The neighborhood size $k$ is also
recalibrated using the corresponding value of $q$.
Changing $q$ can therefore alter the active-review trajectory,
rather than only the final automatic decisions.

\begin{table*}[t]
\centering
\caption{
\textbf{Sensitivity to similarity-floor quantile $q$.}
Deferred, Automatic FG Recall, and False Clear Rate are percentages;
Conditional FG Macro-F1 is reported on a 0--1 scale.
Reviews denotes the number of active reviews before termination.
Values are five-seed mean $\pm$ standard deviation.
The canonical setting is $q=0$.
}
\label{tab:supp_q_sensitivity}
\footnotesize
\setlength{\tabcolsep}{4pt}
\begin{tabular}{llccccc}
\toprule
Dataset & $q$ &
\makecell{Deferred\\(\%) $\downarrow$} &
\makecell{Auto-FG\\Recall (\%) $\uparrow$} &
\makecell{Cond. FG\\Macro-F1 $\uparrow$} &
\makecell{False Clear\\Rate (\%) $\downarrow$} &
\makecell{Active\\Reviews} \\
\midrule
\multirow{4}{*}{ENA24}
& $0$   & $13.0{\pm}1.2$ & $74.1{\pm}0.5$ & $.943{\pm}.010$ & $2.3{\pm}0.7$ & $800{\pm}0$ \\
& $.05$ & $13.8{\pm}0.8$ & $73.8{\pm}0.8$ & $.950{\pm}.010$ & $2.2{\pm}0.7$ & $800{\pm}0$ \\
& $.10$ & $14.2{\pm}1.0$ & $73.6{\pm}0.6$ & $.953{\pm}.004$ & $2.1{\pm}0.6$ & $800{\pm}0$ \\
& $.25$ & $16.4{\pm}0.3$ & $71.7{\pm}0.9$ & $.959{\pm}.005$ & $2.2{\pm}0.7$ & $800{\pm}0$ \\
\midrule
\multirow{4}{*}{LoTE-Animal}
& $0$   & $17.6{\pm}5.0$ & $74.4{\pm}3.2$ & $.920{\pm}.024$ & $4.1{\pm}0.5$ & $800{\pm}0$ \\
& $.05$ & $18.0{\pm}5.6$ & $74.1{\pm}3.3$ & $.923{\pm}.020$ & $4.2{\pm}0.8$ & $800{\pm}0$ \\
& $.10$ & $18.5{\pm}4.5$ & $73.2{\pm}3.7$ & $.930{\pm}.025$ & $4.4{\pm}0.6$ & $800{\pm}0$ \\
& $.25$ & $20.4{\pm}5.0$ & $72.1{\pm}3.8$ & $.934{\pm}.016$ & $4.0{\pm}0.6$ & $800{\pm}0$ \\
\midrule
\multirow{4}{*}{WAUD}
& $0$   & $0.0{\pm}0.1$ & $94.6{\pm}0.5$ & $.949{\pm}.015$ & $1.0{\pm}0.4$ & $711{\pm}104$ \\
& $.05$ & $0.1{\pm}0.1$ & $94.2{\pm}0.4$ & $.950{\pm}.012$ & $1.2{\pm}0.3$ & $758{\pm}37$ \\
& $.10$ & $0.3{\pm}0.3$ & $94.2{\pm}0.3$ & $.955{\pm}.008$ & $1.1{\pm}0.3$ & $788{\pm}25$ \\
& $.25$ & $0.9{\pm}0.8$ & $93.3{\pm}0.4$ & $.953{\pm}.013$ & $1.3{\pm}0.4$ & $785{\pm}34$ \\
\midrule
\multirow{4}{*}{MMLA-Mpala}
& $0$   & $0.0{\pm}0.0$ & $95.6{\pm}0.1$ & $.996{\pm}.001$ & $2.5{\pm}0.1$ & $562{\pm}43$ \\
& $.05$ & $0.0{\pm}0.0$ & $95.4{\pm}0.4$ & $.997{\pm}.001$ & $2.6{\pm}0.4$ & $606{\pm}43$ \\
& $.10$ & $0.0{\pm}0.0$ & $95.4{\pm}0.3$ & $.998{\pm}.001$ & $2.5{\pm}0.3$ & $634{\pm}21$ \\
& $.25$ & $0.0{\pm}0.0$ & $95.2{\pm}0.3$ & $.998{\pm}.001$ & $2.5{\pm}0.3$ & $675{\pm}36$ \\
\bottomrule
\end{tabular}
\end{table*}

Increasing $q$ reduces foreground recovery while improving
conditional recognition on the two camera-trap pools. On ENA24,
Conditional FG Macro-F1 increases from $.943$ to $.959$ between
$q=0$ and $q=.25$, while Automatic FG Recall decreases from
$74.1\%$ to $71.7\%$ and Deferred increases from $13.0\%$
to $16.4\%$. LoTE-Animal exhibits a similar trade-off.

The aerial pools are less sensitive over this range.
MMLA-Mpala reaches zero Deferred workload for every tested
$q$, although higher quantiles require more active reviews
before termination. The canonical $q=0$ favors foreground
recovery rather than maximizing conditional recognition
on the automatically handled subset.

\subsection{Fixed Neighborhood Size}
\label{sec:supp_k_sensitivity}

The canonical method selects $k$ separately for each dataset
and seed using the held-out calibration bank. We instead fix
$k\in\{5,7,9,11\}$ for the entire active-review run.
The foreground and background vote thresholds follow the
same strict-two-thirds rule as the canonical method
(Eq.~\ref{eq:supp_vote_threshold}). Other triage and
acquisition settings remain unchanged.

\begin{table*}[t]
\centering
\caption{
\textbf{Sensitivity to a fixed neighborhood size $k$.}
Each setting uses its corresponding deterministic vote
threshold instead of selecting $k$ from the calibration bank.
Metrics and units follow Table~\ref{tab:supp_q_sensitivity}.
Values are five-seed mean $\pm$ standard deviation.
}
\label{tab:supp_k_sensitivity}
\footnotesize
\setlength{\tabcolsep}{4pt}
\begin{tabular}{llccccc}
\toprule
Dataset & $k$ &
\makecell{Deferred\\(\%) $\downarrow$} &
\makecell{Auto-FG\\Recall (\%) $\uparrow$} &
\makecell{Cond. FG\\Macro-F1 $\uparrow$} &
\makecell{False Clear\\Rate (\%) $\downarrow$} &
\makecell{Active\\Reviews} \\
\midrule
\multirow{4}{*}{ENA24}
& 5  & $13.3{\pm}1.0$ & $74.1{\pm}0.5$ & $.943{\pm}.009$ & $2.1{\pm}0.4$ & $800{\pm}0$ \\
& 7  & $13.2{\pm}0.9$ & $73.8{\pm}0.5$ & $.953{\pm}.006$ & $2.7{\pm}0.4$ & $800{\pm}0$ \\
& 9  & $22.9{\pm}1.1$ & $67.0{\pm}0.9$ & $.970{\pm}.006$ & $1.4{\pm}0.3$ & $800{\pm}0$ \\
& 11 & $22.6{\pm}0.8$ & $66.8{\pm}0.6$ & $.970{\pm}.003$ & $1.8{\pm}0.2$ & $800{\pm}0$ \\
\midrule
\multirow{4}{*}{LoTE-Animal}
& 5  & $15.1{\pm}0.7$ & $76.1{\pm}0.9$ & $.915{\pm}.008$ & $4.3{\pm}0.5$ & $800{\pm}0$ \\
& 7  & $14.6{\pm}0.8$ & $75.8{\pm}0.7$ & $.904{\pm}.006$ & $4.7{\pm}0.9$ & $800{\pm}0$ \\
& 9  & $26.5{\pm}0.8$ & $68.7{\pm}0.6$ & $.960{\pm}.005$ & $3.1{\pm}0.3$ & $800{\pm}0$ \\
& 11 & $26.3{\pm}0.7$ & $69.0{\pm}0.5$ & $.962{\pm}.002$ & $3.1{\pm}0.6$ & $800{\pm}0$ \\
\midrule
\multirow{4}{*}{WAUD}
& 5  & $0.1{\pm}0.1$ & $94.6{\pm}0.5$ & $.956{\pm}.006$ & $1.1{\pm}0.4$ & $779{\pm}30$ \\
& 7  & $0.0{\pm}0.0$ & $94.5{\pm}0.3$ & $.945{\pm}.013$ & $1.1{\pm}0.3$ & $685{\pm}93$ \\
& 9  & $3.2{\pm}0.7$ & $92.6{\pm}0.6$ & $.964{\pm}.007$ & $1.1{\pm}0.3$ & $800{\pm}0$ \\
& 11 & $1.6{\pm}1.0$ & $93.3{\pm}0.5$ & $.948{\pm}.015$ & $0.9{\pm}0.1$ & $800{\pm}0$ \\
\midrule
\multirow{4}{*}{MMLA-Mpala}
& 5  & $0.0{\pm}0.0$ & $95.6{\pm}0.3$ & $.997{\pm}.000$ & $2.4{\pm}0.2$ & $623{\pm}50$ \\
& 7  & $0.0{\pm}0.0$ & $95.5{\pm}0.1$ & $.996{\pm}.001$ & $2.6{\pm}0.1$ & $559{\pm}39$ \\
& 9  & $0.4{\pm}0.2$ & $95.0{\pm}0.1$ & $.999{\pm}.000$ & $2.1{\pm}0.1$ & $800{\pm}0$ \\
& 11 & $0.1{\pm}0.1$ & $95.0{\pm}0.5$ & $.998{\pm}.000$ & $2.4{\pm}0.5$ & $800{\pm}0$ \\
\bottomrule
\end{tabular}
\end{table*}

The fixed-$k$ results show distinct operating points.
On ENA24 and LoTE-Animal, $k=5$ or $7$ produces greater
foreground recovery and fewer Deferred proposals than
$k=9$ or $11$. Larger neighborhoods improve Conditional
FG Macro-F1 but leave more proposals for expert review.
On ENA24, Deferred increases from approximately $13\%$
to $23\%$ between these groups.

The aerial datasets also favor smaller neighborhoods for
reducing Deferred workload. On MMLA-Mpala, $k=5$ and $7$
exhaust the Deferred pool before the review budget is
reached, while $k=9$ and $11$ use all 800 reviews.

\subsection{Additional Parameter Checks}
\label{sec:supp_additional_sensitivity}

\paragraph{Calibration-bank size.}
We also evaluate calibration banks of 25, 50, 100, and
200 proposals, recalibrating $k$ for each setting. The
alternate banks are sampled independently rather than as
nested subsets. Because calibration proposals are excluded
from acquisition and evaluation, these experiments also
change the active and evaluation pools.

There is no consistent improvement from increasing the
bank size. On ENA24, the 25-, 50-, and 100-proposal
settings have Correct Auto between $73.5\%$ and $74.5\%$,
while the 200-proposal setting falls to $71.2\%$ with
greater seed variation. The results do not isolate the
effect of calibration-label count from changes in the
held-out proposal identities.

\paragraph{Density-estimation neighborhood.}
We vary the number of neighbors used to estimate local
density for DM acquisition over
$K_{\mathrm{dens}}\in\{10,20,30,50\}$, with 30 used
canonically. Density is computed over the initial
acquisition pool and reused as proposals are removed.

LoTE-Animal Correct Auto varies from $71.2\%$ to
$71.8\%$, and Automatic FG Recall from $74.0\%$ to
$74.6\%$. ENA24 exhibits greater variation in Deferred
workload and False Clear Rate, but no consistent
improvement from increasing the density neighborhood.

\paragraph{Density-moderation strength.}
For percentile-ranked density $r$, the unnormalized mathematical
weight is
\begin{equation}
    w_{\alpha}(r)=[r(1-r)]^\alpha,
    \qquad \alpha\in\{0.5,1,2\}.
    \label{eq:supp_alpha_weight}
\end{equation}
The canonical $\alpha=1$ row is taken from the original DM runs.
The $\alpha=0.5$ and $\alpha=2$ reruns use
$(r+10^{-4})^\alpha(1-r+10^{-4})^\alpha$, followed by
per-round mean normalization for numerical stability.

On LoTE-Animal, Correct Auto remains between $71.4\%$
and $71.8\%$, and Automatic FG Recall between $74.2\%$
and $74.4\%$. MMLA-Mpala reaches zero Deferred workload
for all three settings. The tested moderation strengths
produce similar endpoint results on these two pools,
although WAUD shows greater seed variation at $\alpha=2$.

%% file: supp/6_runtime.tex
\section{Runtime Analysis}
\label{sec:supp_runtime}

We measure runtime after proposal generation and frozen feature
extraction, excluding SAM3 inference, BioCLIP2/DINOv3 feature
extraction, and human review. The controlled baseline measurements
use a pinned CPU core, single-threaded execution, an untimed
warm-up, and method-order rotation. For each seed, we report
the median of six repeated runs, followed by mean and standard
deviation across five seeds.

\subsection{Online Runtime}

Table~\ref{tab:supp_online_runtime} separates cumulative online
runtime into acquisition, automatic decisions, and repeated
recalibration where applicable. Symmetric $k$-NN reuses the
DM-acquired support of Prototype Triage, giving both methods
matching acquisition times. Prototype Triage has no separate
recalibration stage in this breakdown; class-specific similarity
floors are updated within the decision procedure.

\begin{table}[t]
\centering
\caption{
\textbf{Controlled online-runtime breakdown (s).}
Mean $\pm$ standard deviation across five seeds.
Recal.\ denotes a separate repeated recalibration stage.
Component values are rounded.
}
\label{tab:supp_online_runtime}
\scriptsize
\setlength{\tabcolsep}{1.8pt}
\renewcommand{\arraystretch}{0.92}
\begin{tabular}{llrrrr}
\toprule
Dataset & Method & Acq. & Decision & Recal. & Total \\
\midrule
\multirow{3}{*}{ENA24}
& Triage & $0.47{\pm}0.01$ & $1.40{\pm}0.03$ & $0$ & $1.873{\pm}0.037$ \\
& $k$-NN & $0.47{\pm}0.01$ & $3.59{\pm}0.06$ & $0.66{\pm}0.02$ & $4.716{\pm}0.052$ \\
& LR     & $0$              & $1.11{\pm}0.83$ & $0.96{\pm}0.63$ & $2.074{\pm}1.460$ \\
\midrule
\multirow{3}{*}{LoTE}
& Triage & $2.17{\pm}0.41$ & $5.20{\pm}1.02$ & $0$ & $7.368{\pm}1.435$ \\
& $k$-NN & $2.17{\pm}0.41$ & $12.46{\pm}0.16$ & $0.46{\pm}0.02$ & $15.092{\pm}0.306$ \\
& LR     & $0.01{\pm}0.01$ & $1.47{\pm}1.14$ & $0.60{\pm}0.41$ & $2.074{\pm}1.538$ \\
\midrule
\multirow{3}{*}{WAUD}
& Triage & $0.35{\pm}0.03$ & $0.54{\pm}0.09$ & $0$ & $0.890{\pm}0.121$ \\
& $k$-NN & $0.35{\pm}0.03$ & $4.79{\pm}0.76$ & $0.26{\pm}0.04$ & $5.400{\pm}0.828$ \\
& LR     & $0$              & $0.09{\pm}0.01$ & $0.05{\pm}0.01$ & $0.147{\pm}0.020$ \\
\midrule
\multirow{3}{*}{MMLA}
& Triage & $0.39{\pm}0.01$ & $0.45{\pm}0.02$ & $0$ & $0.840{\pm}0.025$ \\
& $k$-NN & $0.39{\pm}0.01$ & $5.24{\pm}0.52$ & $0.19{\pm}0.02$ & $5.816{\pm}0.536$ \\
& LR     & $0$              & $0.94{\pm}0.38$ & $0.39{\pm}0.04$ & $1.339{\pm}0.414$ \\
\bottomrule
\end{tabular}
\end{table}

Prototype Triage is $2.0$--$6.9\times$ faster than Symmetric
$k$-NN in cumulative online runtime. The difference comes mainly
from lower repeated decision cost and the absence of a separate
recalibration stage. LR is faster on LoTE-Animal and WAUD, but
follows its own uncertainty-sampling trajectory rather than
replaying the same reviewed support.

\subsection{Acquisition Runtime}

Table~\ref{tab:supp_acquisition_runtime} reports cumulative
acquisition-selector time across active rounds, excluding feature
extraction and triage. Unlike the acquisition column in
Table~\ref{tab:supp_online_runtime}, which includes per-round
loop bookkeeping, these measurements time only the selector call.

\begin{table}[t]
\centering
\caption{
\textbf{Cumulative acquisition-selection time (s).}
Five-seed mean; feature extraction, triage, and surrounding
loop bookkeeping are excluded.
}
\label{tab:supp_acquisition_runtime}
\small
\setlength{\tabcolsep}{2.4pt}
\renewcommand{\arraystretch}{0.92}
\begin{tabular}{lrrrr}
\toprule
Method & ENA24 & LoTE & WAUD & MMLA \\
\midrule
Random     & 0.007 & 0.008 & 0.006 & 0.004 \\
$k$-center & 0.103 & 0.429 & 0.058 & 0.071 \\
TypiClust  & 66.1  & 157.5 & 52.4  & 37.1 \\
DACS-TF    & 11.4  & 23.8  & 7.6   & 8.3 \\
Facility   & 9.6   & 43.0  & 6.4   & 6.8 \\
DW         & 0.064 & 0.221 & 0.052 & 0.051 \\
DM         & 0.073 & 0.288 & 0.069 & 0.077 \\
\bottomrule
\end{tabular}
\end{table}

Random, $k$-center, DW, and DM require less than one second
of cumulative selection time on every dataset. Facility and
DACS-TF are more expensive, while TypiClust has the highest
selection cost. DM requires only $0.069$--$0.288$\,s across
the evaluated pools.

These timings characterize computation after proposal embeddings
have been extracted, not end-to-end deployment latency.

%% file: supp/7_qualitative.tex
\section{Qualitative Examples}
\label{sec:supp_qualitative}

\subsection{Reference-Box Mismatch Under Occlusion}
\label{sec:supp_qual_misalignment}

\begin{figure*}[t]
    \centering
    \includegraphics[width=\textwidth]{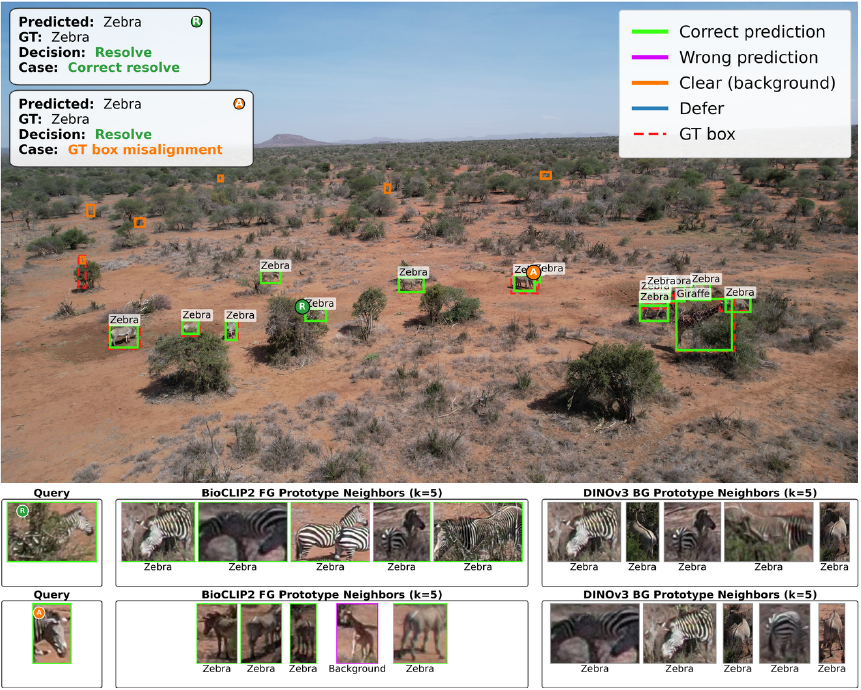}
    \caption{
    Qualitative MMLA-Mpala example illustrating reference-box mismatch under
    occlusion. The upper query is correctly Resolved as zebra. In the lower
    query, SAM3 detects the visible head and upper body of an occluded zebra,
    while the reference box spans a substantially larger extent of the animal.
    The resulting IoU falls below the assignment threshold, causing the
    proposal to be labeled as background for evaluation despite corresponding
    to a genuine animal detection. The retrieved BioCLIP2 foreground neighbors
    also show coherent visual variation within the merged zebra class.
    }
    \label{fig:qual_mmla_misaligned}
\end{figure*}

Figure~\ref{fig:qual_mmla_misaligned} shows an MMLA-Mpala example in which
SAM3 localizes the visible head and upper body of an occluded zebra, while
the reference annotation spans a substantially larger portion of the animal.
The proposal therefore falls below the IoU assignment threshold and is
treated as background by the evaluation protocol, even though it corresponds
to a genuine animal detection. Prototype Triage nevertheless Resolves the
proposal as zebra, illustrating how reference-box mismatch under occlusion
can cause a valid localization to be counted as background during
quantitative evaluation. The retrieved BioCLIP2 neighbors also show coherent
visual variation within the merged zebra class, including examples visually
resembling different zebra types.

\subsection{Confusion Between Visually Similar Species}
\label{sec:supp_qual_confusion}

\begin{figure*}[t]
    \centering
    \includegraphics[width=\textwidth]{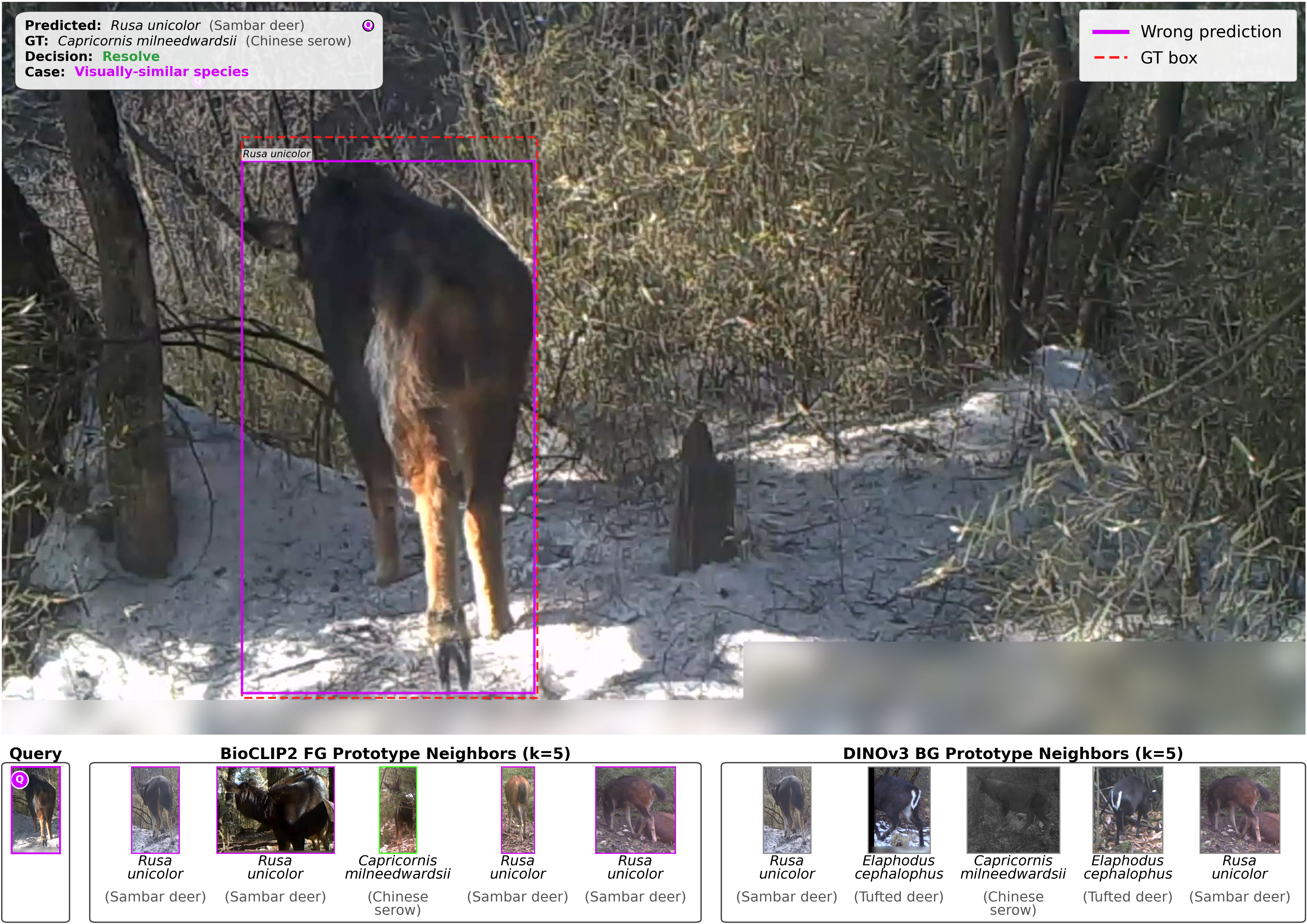}
    \caption{
    Qualitative LoTE-Animal example of confusion between visually similar
    species. The ground-truth animal is a Chinese serow
    (\emph{Capricornis milneedwardsii}), but Prototype Triage Resolves it as
    sambar deer (\emph{Rusa unicolor}). Four of the five nearest BioCLIP2
    foreground prototypes are sambar deer, producing strong but incorrect
    local foreground support. The example illustrates how visually similar
    species can remain difficult to separate even when the prototype
    neighborhood is internally coherent.
    }
    \label{fig:qual_lote_confusion}
\end{figure*}

Figure~\ref{fig:qual_lote_confusion} shows a LoTE-Animal example in which
Prototype Triage incorrectly Resolves a Chinese serow
(\emph{Capricornis milneedwardsii}) as sambar deer
(\emph{Rusa unicolor}). Four of the five BioCLIP2 foreground neighbors are
sambar deer prototypes, while only one supports the ground-truth class.
The query is therefore surrounded by locally coherent but incorrect
foreground evidence.

\subsection{Missed Reference Annotation}
\label{sec:supp_qual_missing_gt}

\begin{figure*}[t]
    \centering
    \includegraphics[width=\textwidth]{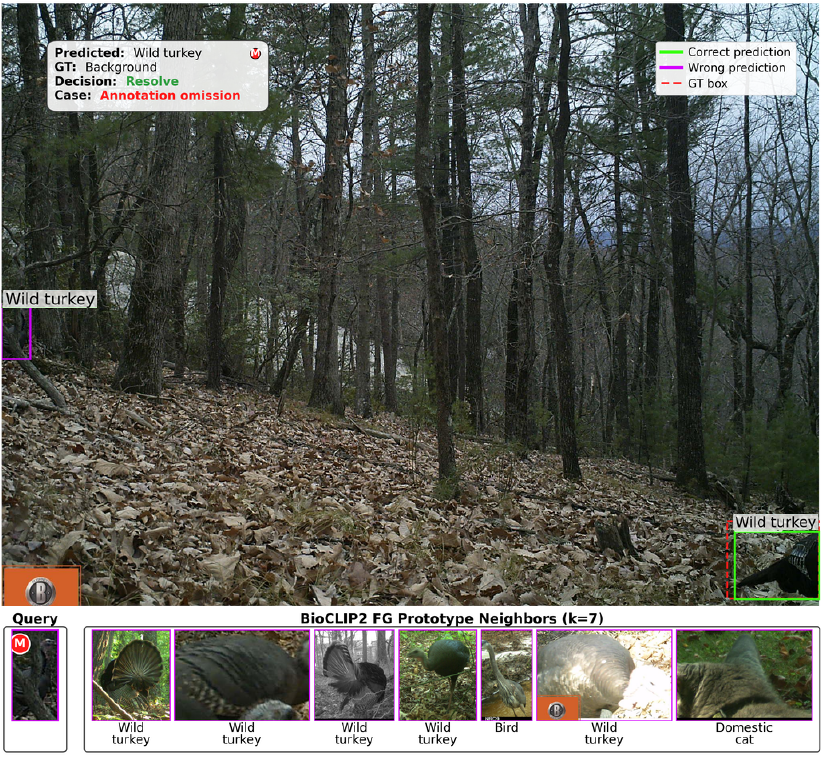}
    \caption{
    Qualitative ENA24 example of a missed reference annotation. The wild
    turkey is difficult to distinguish visually from the surrounding forest
    and is absent from the reference annotations. SAM3 nevertheless localizes
    the animal, and Prototype Triage Resolves the proposal as wild turkey,
    with most of its nearest BioCLIP2 foreground prototypes also belonging to
    the wild-turkey class. Since no corresponding reference box is available,
    the proposal is treated as background by the quantitative evaluation
    despite appearing to be a valid animal detection and recognition.
    }
    \label{fig:qual_turkey_missing}
\end{figure*}

Figure~\ref{fig:qual_turkey_missing} shows an ENA24 example in which a wild
turkey is only weakly visible against the forest background and is absent
from the reference annotations. The detector nevertheless localizes the
animal, and Prototype Triage Resolves the proposal as wild turkey using a
foreground neighborhood dominated by wild-turkey prototypes. Under the
proposal-assignment protocol, the absence of a matching reference box causes
this valid detection to be counted as background.

